\documentclass[12pt,a4paper]{Thesis} 

\graphicspath{%
	{./Pictures/}%
	{./Figures/}%
}

\let\degree\relax

\usepackage{bm}
\usepackage{adjustbox}
\usepackage{amsmath}
\usepackage[amssymb]{SIunits}
\usepackage{amssymb}
\usepackage{multirow}
\usepackage{wrapfig}
\usepackage{enumitem}
\usepackage{subcaption}
\usepackage{mathtools}
\usepackage{diagbox}
\usepackage{lipsum}
\usepackage[ruled,vlined,noresetcount]{algorithm2e}
\usepackage[square, numbers, comma, sort&compress]{natbib} 

\title{\ttitle} 

\usepackage{float}
\usepackage{color,soul}               
\usepackage{enumerate}
\usepackage{Styles/mydefs}

\begin{document}

\frontmatter 

\setstretch{1.3} 

\pagestyle{plain} 



\maketitle



\setstretch{1.3} 

\acknowledgements{\addtocontents{toc}{\vspace{0.8em}} 
First and foremost, I wish to express my greatest gratitude to my advisor Prof. Erik Cambria, for his valuable support during the course of my study and research. He has advised me ever since I wrote my first machine learning model. His experience and passion have encouraged me in all the time of my academic research.

Apart from my advisor, I would also like to thank other senior researchers who offered me useful suggestions and support. Most notably, Prof. Soujanya Poria, Prof. Frank Xing, Prof. Minlie Huang, Prof. Iti Chaturvedi, Dr. Yukun Ma, Dr. Qian Chen, Dr. Qian Liu, Dr. Rui Mao, Dr. Yang Li, Dr. Haiyun Peng, and Dr. Xiaoshi Zhong often let me pick their mind.

My labmates and collaborators, such as Mr. Jinjie Ni, Mr. Vlad Pandelea, Ms. Luyao Zhu, Mr. Wei Li, Dr. Hao Zhou and Dr. Devmanyu Hararika were always there to lend me a hand. And I often learned new things from discussions with them.

Administrative staff from the SCSE graduate office, including Mr. Ker Shen Hui, Ms. Chiam Poh Ling, Ms. Len Ah Chan and Ms. Juliana Binte Jaapar helped me navigate many matters.

I want to thank all my colleagues and friends in Singapore for their wonderful company.

Finally, my gratitude and admiration go to my parents. They have been a consistent support to me and their lifelong learning philosophy inspires me greatly.
}


\pagestyle{plain} 
\emph{``What if solving one problem could lead to solutions to thousands more?''}

\begin{flushright}
---DeepMind
\end{flushright}
\null\vfill 

\begin{center}
\large{To my dear family}
\end{center}
\vfill\vfill\null 
\cleardoublepage 

\pagestyle{plain} 

\tableofcontents 

\addtotoc{Summary} 

\Summary{\addtocontents{toc}{\vspace{0.8em}}}


The goal of building dialogue agents that can converse with humans naturally has been a long-standing dream of researchers since the early days of artificial intelligence. The well-known Turing Test proposed to judge the ultimate validity of an artificial intelligence agent on the indistinguishability of its dialogues from humans'. It should come as no surprise that human-level dialogue systems are very challenging to build. But, while early effort on rule-based systems found limited success, the emergence of deep learning enabled great advance on this topic.

In this thesis, we focus on methods that address the numerous issues that have been imposing the gap between artificial conversational agents and human-level interlocutors. These methods were proposed and experimented with in ways that were inspired by general state-of-the-art AI methodologies. But they also targeted the characteristics that dialogue systems possess.

First of all, we expand the variety of information that dialogue systems can be dependent on. In its simplest and most common form, a dialogue consists of responses and their preceding textual context. This representation, however, falls short compared to real-world human conversation, which is often dependent on other modalities and specific knowledge bases. 

To the end of conditioning dialogues on more modalities, we explore dialogue generation augmented by the audio representation of the input. We design an auxiliary response classification task to learn suitable audio representation for our dialogue generation objective. We use word-level modality fusion for integrating audio features into the Sequence to Sequence learning framework. Our model can generate appropriate responses corresponding to the emotion and emphasis expressed in the audio.

Commonsense knowledge has to be integrated into the dialogue system effectively for it to respond to human utterances in an interesting and engaging way. As the first attempt to integrating a large commonsense knowledge base into end-to-end conversational models, we propose a model to jointly take into account the context and its related commonsense knowledge for selecting an appropriate response. We demonstrate that the knowledge-augmented models are superior to their knowledge-free counterparts.

While the two directions mentioned above endeavor to ground the dialogues on various new information, they are not the only challenges that dialogue systems face. Traditionally, The goal of building intelligent dialogue systems has largely been separately pursued under two paradigms: task-oriented dialogue systems, which perform task-specific functions, and open-domain dialogue systems, which focus on non-goal-oriented chitchat. The two dialogue modes can potentially be intertwined together seamlessly in the same conversation, as easily done by a friendly human assistant. This thesis also covers our effort on addressing the problem of fusing the two dialogue modes in multi-turn dialogues. We build a new dataset FusedChat, which contains conversation sessions containing exchanges from both dialogue modes with inter-mode contextual dependency. We propose two baseline models on this task and analyze their accuracy.

Last but not least, we demonstrate our effort on addressing the computational efficiency issue that large-scale retrieval-based dialogue systems face. Strong retrieval-based dialogue systems that are based on a large natural candidate set can produce diverse and controllable responses. However, a large candidate set could be computationally costly. We propose methods that support a fast and accurate response retrieval system. To boost accuracy, we adopt a knowledge distillation approach where a very strong yet computationally expensive joint encoding model is used to facilitate training our encoders. We then boost the retrieval speed by adopting a learning-based candidate screening method to further reduce inference time. We demonstrate that our model performs strongly in terms of retrieval accuracy and speed trade-off.

In summary, this thesis systematically demonstrates our effort on innovating dialogue systems. We believe that the research questions that we focus on are important aspects for ultimately improving automated dialogue agents to human-level. With our effort of innovating dialogue systems spanning the last 4 years, and state-of-the-art NLP models fast evolving year by year, we note that the models used in some of our works in the earlier years (e.g., LSTMs) cannot compete with the state-of-the-art models available today (e.g., GPT3). In such cases, we briefly and systematically explain following works (current state-of-the-art) that stemmed from the methodologies shown in our work.


\pagestyle{plain} 

\listoffigures 

\listoftables 


%
%


%
%


\cleardoublepage  

\listofnomenclature{ll} 
{
\multicolumn{2}{l}{}
\\
Seq2Seq                      & Sequence to Sequence\\
TOD                     & Task-oriented Dialogue\\
ODD                     & Open-domain Dialogue\\
LSTM                      & Long-short Term Memory\\
GPT                      & Generative Pretrained Transformer (117M parameters)\\
GPT2                     & Generative Pretrained Transformer (2nd version, 1.5B parameters)\\
GPT3                     & Generative Pretrained Transformer (3rd version, 175B parameters)\\
NLP          & Natural Language Processing\\
BERT           & Bidirectional Encoder Representations from Transformers\\
LM             & Language Modeling\\
MIPS             & Maximum Inner Product Search\\
RQ             & Research Question\\
SOTA           & State-of-the-art \\
resp. & respectively \\
a.k.a & also known as \\
e.g. & exemplum gratia (en: for example) \\
et al. & et alia (en: and others) \\
i.e. & id est (en: that is) \\
}


\mainmatter       
\pagenumbering{arabic}
\setstretch{1.3}  


\pagenumbering{arabic}

\pagenumbering{arabic}
\chapter{Introduction} 
\chaptermark{Introduction}
\label{ch:introduction} 


Dialogue systems are becoming increasingly useful in today's world. In contrast to early rigid rule-based automated dialogue systems used in phone calls to customer services, the modern dialogue systems can converse on diverse topics ranging from your pet to recent financial news, and, even further, they can book a restaurant for your birthday party. At present, dialogue systems are one of the hot topics in natural language processing (NLP) and are demanded for both businesses and individual users. The global chatbot market is foreseen to expand from \$3.6 billion in 2020 to \$12.4 billion by 2026\cite{outgrow, ni2021recent}. About 30\% of people use virtual assistants like Google Assistant and Amazon Echo at least once a month as of 2021 \cite{statista}. And it is estimated that about 80\% of all companies implement at least one chatbot somewhere in their business \cite{smallbizgenius}. Such prevalent usages of dialogue systems either drastically reduce repetitive human labor for companies or make the lives of individual users more convenient. 

In the early years, dialogue systems were mostly rule-based (e.g., ELIZA \cite{weizenbaum1966eliza} and PARRY \cite{colby1971artificial}). The general principle for such methods is looking for keywords within a conversation that can be used to instruct an agent to provide predefined responses.  It is important to consider that even though the agent has a script of meaningful and intelligent responses, it has a severely limited understanding of the language itself. The dialogue flows of these systems are predetermined, which restrains the applications of the dialogue systems within very limited scenarios \cite{ni2021recent}.


Similar to other fields in NLP, recent breakthroughs made in deep learning and self-supervised pretraining have greatly propelled the field of dialogue systems forward. Chronologically, two significant paradigm shifts happened recently. The first one was the introduction of neural networks for sequence to sequence (Seq2Seq) learning problems \cite{sutskever2014sequence}. Coupled with abundant data and computational resources, this paradigm first made breakthrough in neural translation and quickly penetrated state-of-the-art dialogue systems \cite{vinyals2015neural}. This paradigm discards feature engineering all together and let the neural network automatically learn to project the conversational context to the response through encoding implicit features and decoding from them. It proved to be able to generate quite natural-sounding responses.

The second paradigm shift coupled large-scale self-supervised pretraining with large neural networks \cite{devlin2018bert, radford2019language}, such as XLNet \cite{yang2019xlnet}, BERT \cite{devlin2018bert}, RoBERTa \cite{liu2019roberta} and GPT3 \cite{brown2020language}, to give them general language understanding capabilities before fine-tuning them on specific tasks. It has been shown that utilizing such pretrained models achieves state-of-the-art results in multiple NLP tasks \cite{devlin2018bert}. Dialogue systems are no exception. A lot of dialogue datasets that people normally train their models on can be relatively small and it is hard to learn enough commonsense knowledge or language variations. Therefore they do not always generalize well to unfamiliar contexts. By initializing with pretrained models, these issues are alleviated and the performance of the dialogue systems is further improved.






In the rest of this chapter, we first illustrate the task definitions we use in this thesis, from various angles including utility, modality and response production method. We then illustrate the backbone models that our proposed methods are based upon. We end with the overall organization of the thesis.

\section{Task Definitions}

Simply put, a dialogue system is a model that maps a conversational context $C$ to a response $R$. There are various types of dialogue systems depending on the utility, modality, response production method. The main content chapters of this thesis focus on various types of dialogue systems. Therefore, to begin with, we illustrate their definitions here.

\subsection{Utility}

Utility, or dialogue mode, refers to the function that a dialogue system serves. They can be classified into task-oriented dialogue (TOD) systems and open-domain dialogue (ODD) systems. The former serve the utility of performing task-specific functions, while the latter focus on non-goal-oriented chitchat.

ODDs generate the response based on any open-domain context and exhibit general chitchat ability. Their primary goal in a conversation is to keep the user engaged and chat over random open-domain topics that he is interested in. For example, Apple's Siri can chat about your day with you. The dialogues can be sustained by commonsense and empathy without the need for any special databases. TOD models are vastly different. The dialogues exist for the purpose of serving specific functions, such as finding restaurants and booking airlines. They operate on closed domains that are often supported by structured databases and APIs [19, 22]. 

Under the scope of this thesis, three characteristics distinguish TODs from ODDs: 

(1) TODs have an entity-centered database. This means the the response $R$ is dependent on the database $D$ in additional to the conversational context $C$. For example, a restaurant reservation bot is dependent on a database of restaurants, containing information such as their names, offerings, etc. 

(2) TODs explicitly predict the user's intent in order to query the database. For example, in order to retrieve the correct restaurant, the user's exact preference on price range and cuisine type needs to be explicitly inferred.

(3) Since TODs are dependent on knowledge bases, they usually have a pre-defined set of dialogue domains and functions. 

In this thesis, ``inter-mode'' or ``multi-mode'' dialogue systems refer to models that can fuse both utilities in the same conversation session. The model's response can be task-oriented or chitchat, and so can the context. We cover our effort in building inter-mode dialogue systems in Chapter \ref{ch:fusedchat_chapter}.

\subsection{Modality}

While most dialogue systems assume textual input (sometimes converted from speech) and output, human conversation is inherently multi-modal. Humans understand each other through audio and video signals that are sometimes beyond what text can convey.

In multimodal dialogue systems, the context $C$ and the response $R$ are potentially multimodal. For example, if the user inputs his message using a microphone, $C$ might contain $C_{text}$, which is the textual message, and $C_{audio}$, which is the audio clip. Similarly, $R$ could be a short video clip of a digital avatar, which would naturally contain $R_{audio}$, i.e., the audio clip and $R_{video}$, i.e., the facial expression and the body gesture of the avatar.

Chapter \ref{ch:audio-chapter} explores an approach of conditioning dialogue systems on audio features in the context.

\subsection{Response Production Method}

Depending on how the response $R$ is generated, dialogue systems can be classified into generation-based and retrieval-based. The former models attempt to generate brand-new responses from scratch. The latter models select the most appropriate response from a pre-constructed response candidate set.

In other words, a generation-based dialogue system learns a sequence transduction model $R = f_1(C)$. The model generates a response based solely on the context. A retrieval-based dialogue system learns a scoring model $f_2(C, R)$ that computes the compatibility of a context and a response. In addition, the system has access to a large response repository $\mathcal{R}$ from where it chooses the best response for each context.

Chapter \ref{ch:commonsense_chapter} explores conditioning dialogue systems on a commonsense knowledge base, which is based on a retrieval-based scenario. Chapter \ref{ch:retrieval-chapter} explores approaches for accelerating the inference speed of large-scale retrieval-based dialogue systems.

\section{Backbone Models}

Backbone models such as long short-term memory networks (LSTM) and Transformers have been based upon by many state-of-the-art models in various NLP tasks in the recent years. Our early work in Chapter \ref{ch:audio-chapter} and \ref{ch:commonsense_chapter} is based upon LSTMs. Our later work in Chapter \ref{ch:fusedchat_chapter} and \ref{ch:retrieval-chapter} is based upon transformers.

\subsection{LSTM} \label{sec:LSTM}

As a version of recurrent neural network, an LSTM network \cite{hochreiter1997long} is good at handling long-term dependencies and can be used to map an utterance to hidden states as fixed-size embedding representations, based on which decoding can be done to predict the next tokens.

The $k$th token in an utterance is first embedded into a vector $e_k$ of dimension $d$ using a word embedding matrix. Then, the hidden representation $h_k$ at time step $k$ for the utterance is defined by: 
\begin{eqnarray}
i_k=\sigma(W_i\cdot[h_{k-1},e_k])\\
\nonumber
f_k=\sigma(W_f\cdot[h_{k-1},e_k])\\
\nonumber
o_k=\sigma(W_o\cdot[h_{k-1},e_k])\\
\nonumber
l_k=tanh(W_l\cdot[h_{k-1},e_k])\\
\nonumber
c_k=f_k\cdot{c_{k-1}}+i_k\cdot{l_k}\\
\nonumber
h_k=o_k\cdot{tanh(c_k)}
\label{eq2}
\end{eqnarray}where $W_i$, $W_f$, $W_o$, $W_l$ $\in \mathcal{R}^{D\times(D+d)}$. An input gate, a memory gate and an output gate, denoted as $i_k$, $f_k$ and $o_k$, are used to update cell state $c_k$ and hidden state $h_k$ iteratively. $D$ is the dimension of hidden state $h_k$. $\sigma$ denotes the sigmoid function.

For the mainstream community, deep LSTMs were considered the state-of-the-art models for many NLP tasks up until the rise of transformers. LSTMs were used as backbone models for our work in Chapters \ref{ch:audio-chapter} and \ref{ch:commonsense_chapter}.

\subsection{Transformer} \label{sec:Transformer}

\cite{vaswani2017attention} brought forward a type of neural networks that processes tokens in a sequence in a parallel manner using Self-Attention. 

\begin{figure*}[h]
  \includegraphics[]{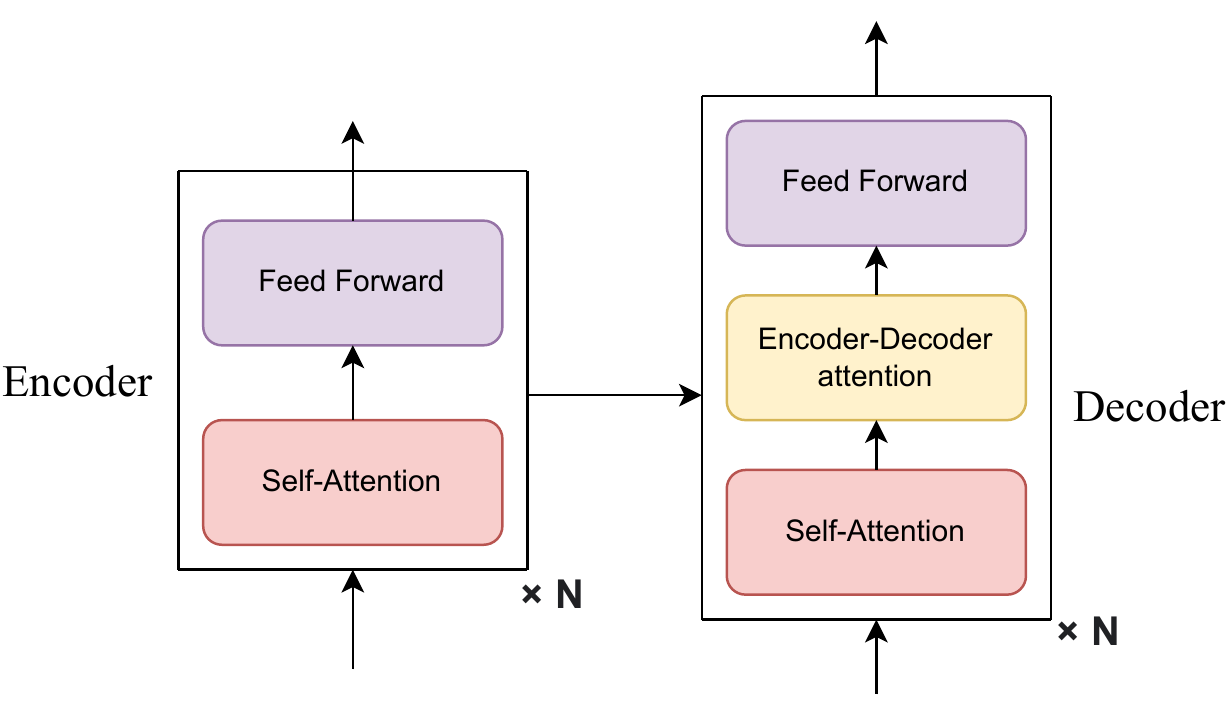}
  \centering
  \caption{Transformers used for Seq2Seq learning as defined in \cite{vaswani2017attention}.}
  \label{fig:memory_module}
\end{figure*}

Self-Attention involves three vectors:

Query: The query is a representation of the current word used to score against all the other words (using their keys).

Key: The key vectors are matched against queries to determine the importance of the words.

Value: Value vectors are ``actual'' word representations. Once how relevant each word is is determined, the values are added up to represent the current word.

We use $Q, K, V$ to represent the query, key, value matrices (vectors from multiple tokens condensed together). They are obtained by multiplying the input matrix $X$ with the transformation matrices $W^Q, W^K, W^V$.

The Self-Attention mechanism can be represented as:

\begin{eqnarray}
Attention(Q, K, V) = softmax(\frac{QK^T}{\sqrt{d_k}})V,
\end{eqnarray}

where $d_k$ is the dimension of the key vector. Instead of performing a single attention function, it is found beneficial to use Multi-Head attention with $h$ parallel attention functions and then concatenate the output.

\begin{eqnarray}
MultiHead(Q, K, V) = Concat(head_1, ..., head_h)W^O\\
head_i = Attention(QW^Q_i, KW^K_i, VW^V_i),  i \in [0, h]
\end{eqnarray}

\cite{vaswani2017attention} used Self-Attention in both the encoder and the decoder in Seq2Seq learning tasks. In the decoder stack, the Encoder-Decoder attention module is used, which is the same as the Self-Attention module except the the query vectors are from the decoder and the key and value vectors are from the encoder.

One key difference between the encoder and the decoder is attention masking. In the encoder, the encoded sequence is fully given and visible, therefore each token can attend to every other token to the left and to the right. In the decoder, however, the training objective is to predict the following tokens given previous tokens. Therefore each token can attend to the tokens to the left.

Transformers are used in most state-of-the-art NLP models at the moment. For example, when used for large-scale pretraining, they become powerful multi-task learners \cite{radford2019language}, as demonstrated in the next two sections.


\subsection{BERT} \label{sec:BERT}

The Bidirectional Encoder Representations from Transformers (BERT, \cite{devlin2018bert}) model is designed to pre-train bidirectional representations from raw text by conditioning on bidirectional context. The pre-trained BERT model can be fine-tuned to create state-of-the-art models for a wide range of tasks, such as question answering and natural language inference, without substantial task-specific architecture modifications.

BERT creates powerful representations for words and sentences that are useful for sentence-level and token-level tasks that be regarded as classification tasks. It is not explicitly created to generate sequences and therefore BERT only contains the encoder part as mentioned in the last section.

BERT is pretrained on two tasks: (1) the Masked Language Modeling task which predicts the masked token based on bidirectional context, and (2) the Next Sentence Prediction task which predicts if one sentence naturally follows another.

The first token of every sequence is always a special classification token ([CLS]). The final hidden state corresponding to this token is used as the aggregated sequence representation for downstream tasks. BERT can naturally generate representation for a single sentence or a sentence pair. For example, a BERT model can be used to encode a context and a response separately. Afterwards a scoring model can use the resulting two representations to calculate their compatibility, through, e.g., inner-product. Alternatively, a BERT model can be used to encode the concatenation of a context and a response as one sequence. The single resulting representation can be used to score how plausible the exchange is, through, e.g., a feed-forward neural network.

In Chapter \ref{ch:retrieval-chapter}, we explore methods that accelerate retrieval-based dialogue systems based on BERT representations.


\subsection{GPT} \label{sec:GPT}

Generative Pretrained Transformers (GPT) are another class of models that proved the effectiveness of large-scale pretraining. In particular, they excel at language generation tasks.

Unlike BERT, which seeks to learn bi-directional representations of words and sentences, GPTs are unidirectional language models which simply learns to predict the next token given previous tokens. GPTs can be used to generate sequences in a straightforward manner using beam search. To achieve this property, GPTs use the decoder architecture as mentioned in Section \ref{sec:Transformer} without Encoder-Decoder
attention, since no input and output sequences are defined during pretraining.

In Chapter \ref{ch:fusedchat_chapter}, we utilise GPTs as the backbone for end-to-end TOD models and inter-mode dialogue models.










\section{Organization}

\begin{figure}[htbp]
  \centering
    \includegraphics[width=0.85\textwidth]{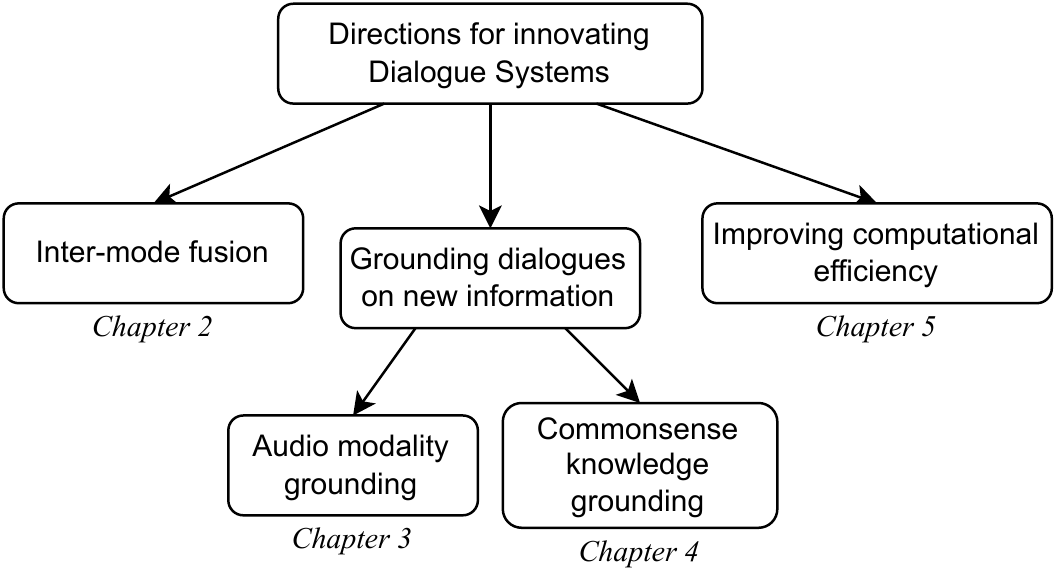}
  \caption{Thesis structure diagram}
  \label{fig:thesis_structure_diagram}
\end{figure}

In this section, we provide an outline of the rest of the thesis. The primary contributions of this thesis are the four main content chapters: Chapters 
\ref{ch:fusedchat_chapter}-\ref{ch:retrieval-chapter} (See the overall concept structure in Figure \ref{fig:thesis_structure_diagram}). The outlines of chapters are as follows:

Chapter \ref{ch:fusedchat_chapter}: Bridging the gap between task-oriented and open-domain dialogues

In this chapter, we illustrate our effort on building dialogue systems that seamless fuse the utilities of performing task-oriented dialogues and open-domain chitchat. We detail our effort on expanding the frequently used MultiWOZ dataset into the new FusedChat Dataset - a new dialogue dataset on which inter-mode dialogue systems can be tested. We also evaluate two baseline approaches on this new dataset.

Chapter \ref{ch:audio-chapter}: Grounding dialogues on non-textual modalities

This chapter showcases our effort on conditioning the dialogue systems on the audio modality in addition to text. We first propose an auxiliary response selection task to learn suitable audio representations from raw noisy audio features. We then propose an Audio-Seq2Seq framework which concatenates audio features to traditional word embeddings. Our audio-augmented model outperforms its audio-free counterpart on perplexity, response diversity and human evaluation.

Chapter \ref{ch:commonsense_chapter}: Grounding dialogues on commonsense knowledge

In this chapter, we demonstrate our effort on augmenting dialogue systems with a large structured commonsense knowledge database. In the retrieval-based scenario, we propose the Tri-LSTM model to jointly take into account the context and commonsense for selecting an appropriate response. Experiments suggest that the knowledge-augmented models are superior to their knowledge-free counterparts in terms of retrieval accuracy.

Chapter \ref{ch:retrieval-chapter}: Improving the computational efficiency of large-scale response retrieval

Our last main content chapter targets the computational efficiency problem. In this chapter, we present methods to improve the inference speed of a large-scale response retrieval model. We used knowledge distillation to leverage the learning power of a cumbersome joint encoding model to improve the performance of our fast individual encoders. Furthermore, to better handle large response candidate sets, we propose a learning-based screening model that makes the retrieval process about 5 times faster with very little accuracy loss. Finally, we demonstrate a pipeline that performs strongly in terms of speed and quality trade-off compared to other retrieval-based models.

Chapter \ref{ch:Conclusion}: Summary

We conclude this thesis by summarizing our contributions, analyzing the role of dialogue systems in AI, and conjecturing future directions.

\chapter{Bridging the gap between task-oriented and open-domain dialogues} 
\chaptermark{Bridging the gap between task-oriented and open-domain dialogues}  
\label{ch:fusedchat_chapter}
The goal of building intelligent dialogue systems has largely been \textit{separately} pursued under two paradigms: task-oriented dialogue (TOD) systems, which perform task-specific functions, and open-domain dialogue (ODD) systems, which focus on non-goal-oriented chitchat. The two dialogue modes can potentially be intertwined together seamlessly in the same conversation, as easily done by a friendly human assistant. Such ability is desirable in conversational agents, as the integration makes them more accessible and useful. This chapter addresses the problem of fusing TODs and ODDs in multi-turn dialogues. Based on the popular TOD dataset MultiWOZ, we build a new dataset FusedChat, by rewriting the existing TOD turns and adding new ODD turns. This procedure constructs conversation sessions containing exchanges from both dialogue modes. It features inter-mode contextual dependency, i.e., the dialogue turns from the two modes depend on each other. Rich dependency patterns such as co-reference and ellipsis are included. The new dataset, with 60k new human-written ODD turns and 5k re-written TOD turns, offers a benchmark to test a dialogue model's ability to perform inter-mode conversations. This is a more challenging task since the model has to determine the appropriate dialogue mode and generate the response based on the inter-mode context. But such models would better mimic human-level conversation capabilities. We propose and evaluate two models on this task, including the \textit{classification-based} two-stage models and the \textit{two-in-one} fused models. 


\section{Introduction}

\begin{figure}[h]
\centering
\includegraphics[width=0.7\linewidth]{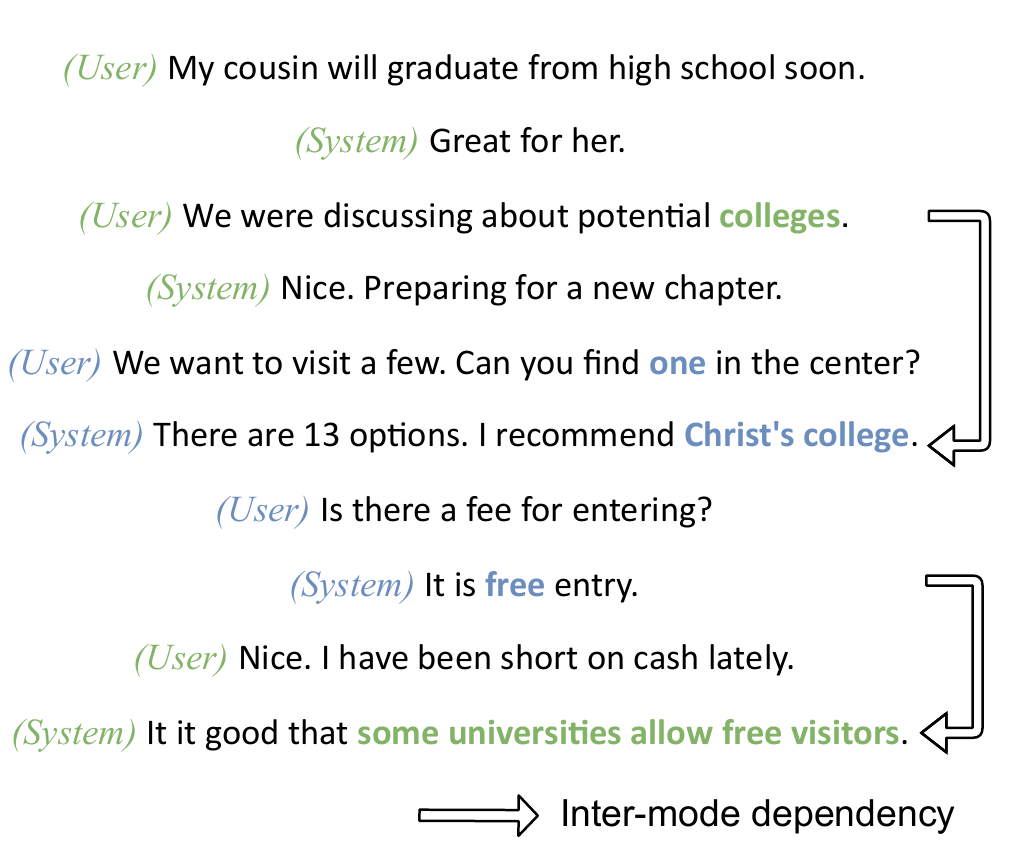}
\caption{Example of interaction with our dialogue system. The conversation between a user and a digital assistant seamlessly interchanges between TOD and ODD modes with strong inter-mode dependency. The conversation involves querying about a college entrance fee (TOD, lines 5 to 8) and chitchat about personal development and finance (ODD, the other lines).}
\label{fig:toy_example}
\end{figure}

According to their utility, two mainstream types of dialogue models can be categorized as ODD models~\cite{adiwardana2020towards,roller2020recipes,zhang2019dialogpt} and the TOD models~\cite{ham2020end,budzianowski2018multiwoz}. ODD models generate the response based on the context and exhibit general chitchat ability. Their primary goal in a conversation is to keep the user engaged and chat over random open-domain topics that he is interested in. The dialogues can be sustained by commonsense without the need for any special databases. TOD models are vastly different. The dialogues exist for the purpose of serving specific functions, such as finding restaurants and booking airlines. They operate on closed domains that are often supported by structured databases and APIs~\cite{budzianowski2018multiwoz,rastogi2020towards}. Commonly three characteristics distinguish them from ODD models: (1) an entity-centered database, (2) explicit dialogue state modeling, and (3) a pre-defined set of dialogue domains and functions (dialogue acts). Humans are able to effortlessly conduct both types of conversations seamlessly together. It is ideal for a dialogue system to be able to do so, because such integration offers a fused system with increased usability. Furthermore, it allows rich interactions between the two dialogue modes, which can not be modeled in either mode independently. Such a dialogue model would better mimic human-level conversation capabilities, e.g., chatting with a friendly assistant (Fig.~\ref{fig:toy_example}).

Despite that numerous datasets have been created in recent years for both ODDs and TODs, there is no high-quality human-written dataset on their fusion, especially with inter-mode contextual dependency. Our work aims to fill this void. We use the popular TOD dataset MultiWOZ~\cite{budzianowski2018multiwoz} as the backbone and let human creators add ODD turns before or after the existing TOD turns. For roughly half the MultiWOZ dialogues, we prepend ODD turns, creating ODD + TOD sessions. For the other half, we append ODD turns, creating TOD + ODD sessions. In both cases, the creator writes an ODD that is contextually related to the existing TOD. We enforce inter-mode dependency in FusedChat. In the prepending case, we make sure the TOD is dependent on the ODD by rewriting the first turn of the TOD, typically with co-reference or ellipsis. In the appending cases, we make sure at least one exchange in the ODD is dependent on concepts or knowledge found in the TOD. In a nutshell, these dependency patterns in our dataset mean that when a dialogue model handles a turn of one dialogue mode, it sometimes has to refer to the contextual information given in the history turns of the other dialogue mode.\\

This new dataset offers a unique test-bed for training and evaluating inter-mode dialogue systems that possess both TOD and ODD capabilities. Traditional dialogue evaluation metrics for both dialogue modes can be used together for inter-mode evaluation. 

We develop and evaluate two baseline models for this new setting: (1) The \textit{classification-based} model. Two response generation models $\mathcal{M}_{tod}$ and $\mathcal{M}_{odd}$ are independently trained on the turns of the respective modes. They generate the response of their respective mode given a conversational context. A separate mode classification model $\mathcal{C}$ is trained and used to determine which mode to invoke given the context. (2) The \textit{two-in-one} fused dialogue model that is trained on dialogue turns of both modes together. It generates a response given any conversational context, by implicitly predicting the dialogue mode as part of sequence generation.

In summary, the main contributions\footnote{\url{https://github.com/tomyoung903/FusedChat}} that this chapter covers are: (1) A new dialogue dataset named FusedChat that fuses TODs and ODDs in multi-turn dialogues. The dialogues feature inter-mode contextual dependency for seamless mode fusion, allowing the dialogue model to better mimic human-level conversation capabilities. FusedChat, with 60k new human-written ODD turns and 5k re-written TOD turns, serves as a new benchmark for inter-mode dialogue systems. Traditional metrics used to gauge TOD and ODD systems separately can be combined to evaluate inter-mode dialogue systems. (2) \textit{Two-in-one} models and \textit{classification-based} models are developed and evaluated as inter-mode dialogue models. Our preliminary experiments suggest that the models evaluated on FusedChat perform worse than their single-mode counterparts evaluated on single-mode datasets. And the more computationally expensive \textit{classification-based} model outperforms the cheaper \textit{two-in-one} fused model. This suggests that effectively fusing different dialogue modes is a challenging task for future work.

\section{Proposed Dataset}

\subsection{FusedChat Construction}

To create inter-mode dialogue sessions, our dataset construction process mainly involves having dialogue creators prepend or append self-written ODDs to existing TODs. A dialogue creator plays the part of both the user and the dialogue system by himself. This self-dialogue setting~\cite{byrne2019taskmaster} avoids misunderstandings between two human creators and improve the consistency of the created dialogues.

For the existing TODs, the MultiWOZ 2.4 dataset~\cite{ye2021multiwoz} is selected because of its popularity in the literature. MultiWOZ contains TODs in 7 domains, including restaurant, attraction, train, police, hospital, taxi and hotel. The user converses with the dialogue agent for a pre-defined set of functions, such as booking restaurants and locating hospitals. Despite that MultiWOZ was created assuming the user is a tourist~\cite{budzianowski2018multiwoz}, we find that most dialogues themselves do not necessarily reflect a tourist persona and allow flexibly adding ODDs. In our FusedChat setting, the dialogue creators are free to add any ODD that is contextually consistent with the existing TOD.

In the following sections, we first discuss the general requirement we set for the added ODDs. We then explain how prepending and appending ODDs are executed and how inter-mode dependency is enforced, respectively.

\subsubsection{General Requirements for the Added ODDs} 


In this section, we describe the general requirements for the added ODDs for both the prepending and appending cases, as rules for the dialogue creators to follow.

(1) Every creator writes fictitious ODDs for \textit{both} the roles of ``system'' and ``user'', where the ``system'' represents an AI conversational agent that is capable of both friendly open-domain conversation (in the added ODDs) and task-oriented dialogues (in the existing MultiWOZ TODs). And ``user'' represents a human speaker that converses with the AI agent for friendly chitchat and to achieve certain task objectives. 

(2) To ensure the relevance between the existing TOD and the added ODD, we encourage the creators to make the ODD revolve around similar or related topics as in the existing TOD segment, e.g., by talking about the same or related concepts in the TOD. The added ODD turns and the existing TOD turns should connect with each other naturally. There should be strong contextual dependency between the two modes (explained in the next 2 sections).

(3) The created dialogues should adhere to the general characteristics of ODDs as opposed to TODs. They should be casual chitchat exchanges that do not require the ``system'' to perform any specific task-oriented functionalities or provide any task-specific information.

\begin{itemize}
  \item Based on the pilot experiment with a sample of creators, we found that the creators had a tendency to write dialogues that are focused on task-specific functionalities, which are technically TODs instead of ODDs as instructed. This is presumably because of a lack of nuanced understanding of their difference, and the ease of fitting those TODs into the context of existing TODs.\\As an aggressive measure to combat this issue, we deployed a real-time turn-level ODD vs TOD classifier, trained on a combination of three traditional ODD datasets~\cite{zhang2018personalizing, smith2020can, dinan2018wizard} and MultiWOZ. In addition, we outline several pitfalls found in the pilot experiment for the creators to avoid, such as letting the system fabricate information that is beyond commonsense.
\end{itemize}

Next, we describe the details on how appending ODDs (TOD + ODD) and prepending ODDs (ODD + TOD) are executed, and how inter-mode dependency is enforced, respectively.

\subsubsection{Appending ODDs}
In the appending scenario, the dialogue creators append an ODD to a provided TOD sampled from the MultiWOZ dataset. The ODD should naturally \textit{follow} the TOD.

\begin{itemize}
\item
We notice that the dialogues from the original MultiWOZ dataset often end with a ``User: Thank you. System: Goodbye.'' exchange. This exchange effectively \textit{ends} the conversation. For appending ODDs, we heuristically remove such exchanges from the end of the TOD based on dialogue act annotations (dialogue-act:thank-you and dialogue-act:goodbye).
\end{itemize}



\begin{figure}[t]
\centering
\includegraphics[width=0.7\linewidth]{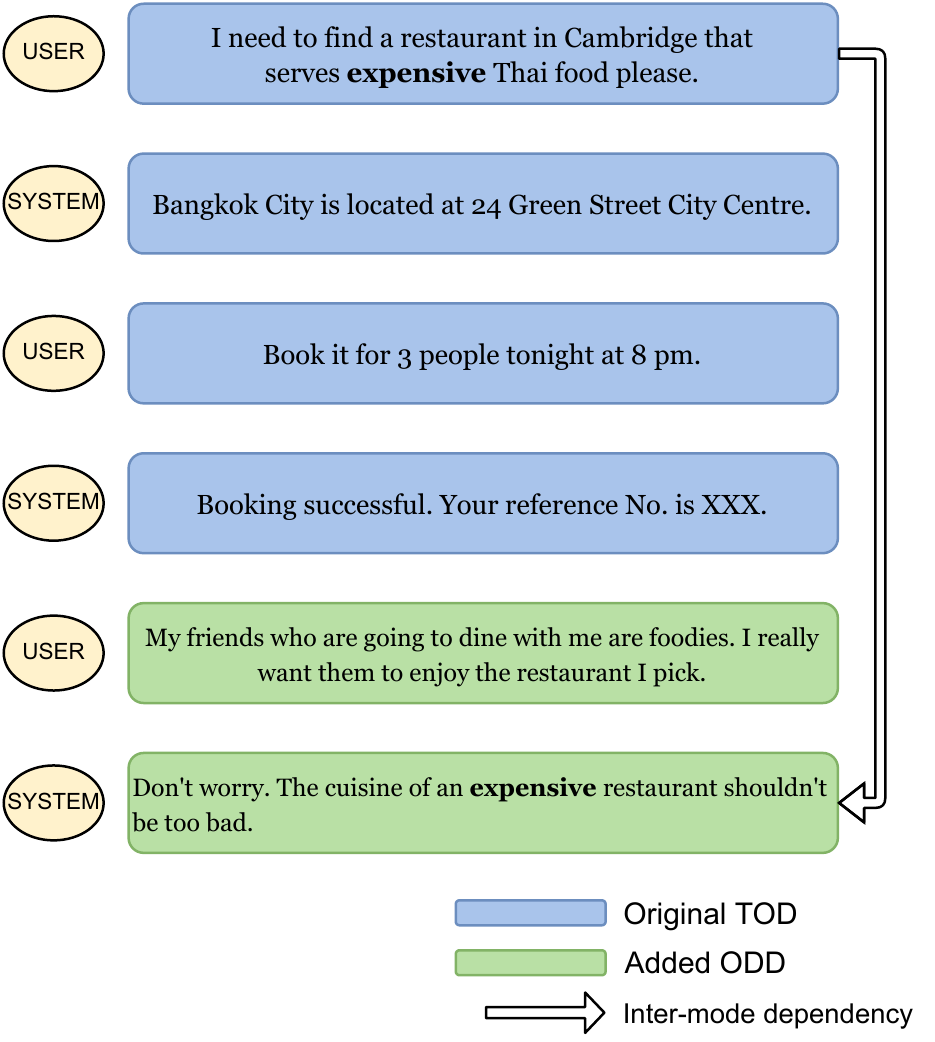}
\caption{An excerpt from a TOD + ODD instance from FusedChat. Note how inter-mode dependency is featured in the last system ODD turn by referring to the concept ``expensive restaurant'' previously mentioned in the TOD.}
\label{fig:dependency_appending}
\end{figure}

In appending cases, the content of the ODD should be dependent on the preceding TOD. We enforce this by letting the creators write at least one round of exchange whose content reflects concepts or knowledge found the existing TOD segment. 

\subsubsection{Prepending ODDs}

In prepending cases, the creator is given a TOD segment from MultiWOZ and asked to prepend an ODD to it. The ODD should naturally \textit{lead to} the provided TOD.

Note that the original TODs in MultiWOZ are self-contained. For our purpose of modeling inter-mode dependency, we conduct utterance rewriting based on co-reference and ellipsis. In FusedChat, they are the key why the TOD is dependent on the prepended ODD.


We want to create ODD + TOD sessions where the TOD is conditioned on the ODD. The key to a successful TOD is dialogue state tracking, where the dialogue system processes the user utterance for [slot type, slot value] pairs (e.g., [Destination: Cambridge]) in order to understand the user's need and respond properly. Our designed method to model inter-mode dependency in our dataset essentially imposes ODD-dependent dialogue state tracking.

\begin{figure}[]
  \centering
  \begin{subfigure}[b]{\linewidth}
  \centering
    \includegraphics[width=0.7\textwidth]{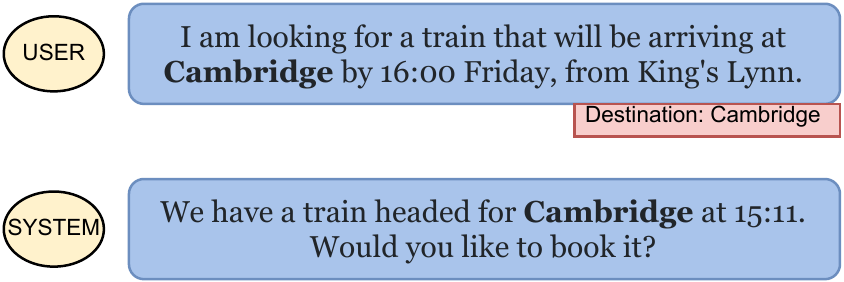}
    \caption{An original TOD exchange with the dialogue state [Destination: Cambridge].}
    \label{fig:original_tod}
  \end{subfigure}
  
  \vspace{0.05\textwidth}%

  \begin{subfigure}[b]{\linewidth}
  \centering
    \includegraphics[width=0.7\textwidth]{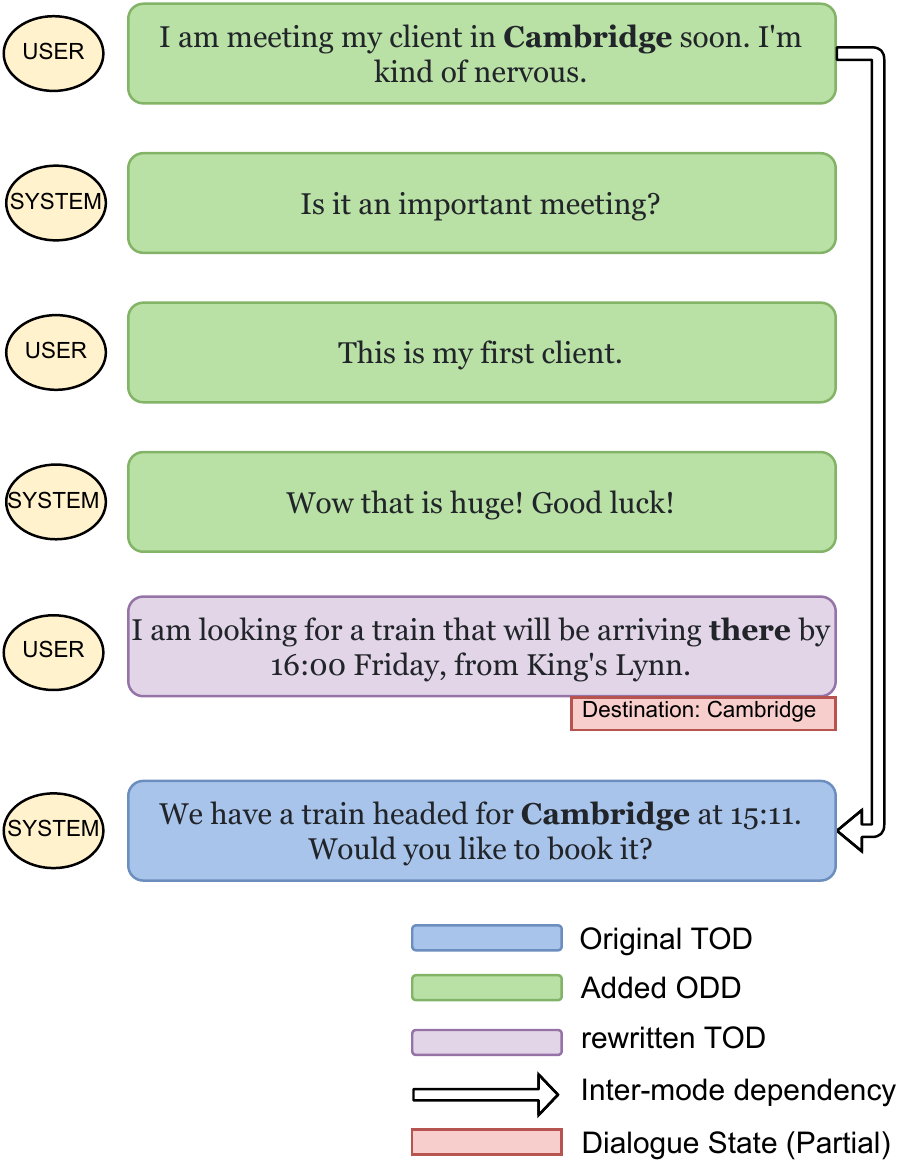}
    \caption{New ODD turns are prepended to the original TOD in (A). Note that the TOD user turn is rewritten. The slot value ``Cambridge'' is mentioned in a prepended ODD turn while co-reference is used in the rewritten user turn. This imposes ODD-dependent dialogue state tracking, forcing the the dialogue system to look for clues in the ODD when it tries to interpret the user's need. }
    \label{fig:}
  \end{subfigure}

\caption{An ODD + TOD instance from FusedChat.}
\label{fig:dependency_prepending}
\end{figure}

We randomly select a slot value mentioned in the first user turn in the TOD, e.g., ``Cambridge'' in Fig.~\ref{fig:dependency_prepending}. We ask the dialogue creators to use the slot value in the prepended ODD, and rewrite the first dialogue user turn accordingly to refer to it implicitly. Rewriting mainly involves co-reference (e.g., ``there'' in Fig.~\ref{fig:dependency_prepending}), and sometimes ellipsis. Co-reference and ellipsis are important features in multi-turn TODs, attracting researchers to sometimes perform special annotations for them in certain TOD datasets~\cite{quan2020risawoz}. See Fig.~\ref{fig:dependency_prepending} for a detailed example on how inter-mode dependency is featured for ODD + TOD sessions.

$ $
\subsection{FusedChat Statistics}

A total of 113 undergraduate students from the authors' university were recruited as dialogue creators for FusedChat. The difference between FusedChat and MultiWOZ mainly lies in the additional ODD turns, either grounding or grounded by the original TODs.
The added ODD turns in FusedChat are a significant extension to the original MultiWOZ dataset. As shown in Table~\ref{tab:statistics}, 60k+ new ODD turns are added, including 8k+ new tokens not present in the original MultiWOZ dataset, significantly expanding the vocabulary.

FusedChat also rewrites the first TOD turns (4670 in total) for the scenario of prepending ODDs. For the scenario appending ODDs, FusedChat discards 11320 TOD turns containing only ``thank-you'' and ``goodbye'' dialogue acts. Table~\ref{tab:partition} shows the training/validation/testing partitions for FusedChat.

\begin{table*}[h]
\centering
\begin{tabular}{|c|c|}
\hline
Total No. turns                                                         & 60579  \\ \hline
Total No. tokens                                               & 680448 \\ \hline
Avg. No. turns per dialogue                                       & 5.81   \\ \hline
Avg. No. tokens per turn                                 & 11.23  \\ \hline
No. unique tokens                                   & 11822  \\ \hline
No. unique tokens not present in MultiWOZ & 8075   \\ \hline
\end{tabular}
\caption{Statistics on the added ODD turns in FusedChat}
\label{tab:statistics}
\end{table*}

\begin{table*}[]
\centering
\begin{tabular}{|c|c|c|c|}
\hline
Partition   & ODD + TOD &  TOD + ODD &   Total \\ \hline
Training    & 3670      &      4768 &      8438         \\ \hline
Validation  & 500       &      500  &      1000           \\ \hline
Testing     & 500       &      500  &      1000          \\ \hline
Total       & 4670      &     5768  &      10438         \\ \hline
\end{tabular}
\caption{FusedChat is composed of ODD + TOD (prepending ODDs) instances and TOD + ODD (appending ODDs) instances.}
\label{tab:partition}
\end{table*}

$ $

\section{Proposed Approaches}\label{sec:approaches}

In this section, we discuss baseline models we developed for inter-mode dialogues.

\subsection{Task Definition}
A multi-turn dialogue system generates a response $R$ based on a multi-turn context $C$. In inter-mode dialogues, $C$ is composed of both TOD and ODD turns. In the FusedChat setting, $R$ can be in either TOD mode or ODD mode, but has to be in only one of the two.z

\subsection{Models} \label{sec:models}
We experiment with two types of models for inter-mode dialogues. (1) The \textit{classification-based} model that is composed of a mode classification model and two response generation models for TOD and ODD separately and (2) the \textit{two-in-one} fused model where a single response generation model can perform both TOD and ODD generation, implicitly determining the response mode.

(1) The \textit{classification-based} model. Two response generation models $\mathcal{M}_{odd}$ and $\mathcal{M}_{tod}$ are independently trained to handle each conversation mode. A separate classification model $\mathcal{C}$ is trained and used to determine which mode of model to invoke given an inter-mode context. Note that all 3 models above take inter-mode context as input.

\begin{itemize}

\item For $\mathcal{M}_{odd}$, we follow~\cite{shuster2019dialogue} and experiment with DialoGPT~\cite{zhang2019dialogpt} as the pretrained model, fine-tuned on all ODD turns in FusedChat. 

\item For $\mathcal{M}_{tod}$, we follow the recent progress on end-to-end modeling for TODs. Dialogue state tracking, dialogue act prediction and response generation have been together cast under a Seq2Seq framework~\cite{hosseini2020simple, ham2020end}. For traditional Seq2Seq-based ODD modeling, the problem is cast as [Context $\rightarrow$ Response]. For Seq2Seq-based TOD modeling, the problem is cast as [Context $\rightarrow$ (Dialogue State, Dialogue Act, Response)], where the three latter parts are concatenated together as one sequence as the generation target. This simplistic form allows TOD models to exploit the benefits of large-scale pretrained models, same as ODD models did. We follow \textit{Neural Pipeline}~\cite{ham2020end} for such a model for $\mathcal{M}_{tod}$, initialized with GPT2 \cite{radford2019language} and fine-tuned on all TOD turns in FusedChat.

\item For $\mathcal{C}$, we follow~\cite{madotto2020adapter} and experiment with BERT~\cite{devlin2018bert} as the pretrained model. The model is fine-tuned on all turns in FusedChat to predict the dialogue mode (TOD vs ODD).

\end{itemize}

(2) The \textit{two-in-one} model. Trained on dialogue turns of both modes, it uses a single model that generates a response given any conversational context by implicitly determining the conversational mode. Similar to~\cite{sun2020adding}, we use an additional $<$ODD$>$ token during sequence construction to indicate when the response is in the ODD mode. The training sequences are composed of [Context $\rightarrow$ ($<$ODD$>$, Response)] and [Context $\rightarrow$ (Dialogue State, Dialogue Act, Response)]. The model is initialized with GPT2 and fine-tuned on all dialogue turns in FusedChat.

For all the models above, best checkpoints for testing are selected based on the full validation set of 1000 instances.

\section{Experiments and Results}\label{sec:fusedchat_benchmark}

\begin{figure}[h]
\centering
\includegraphics[width=0.7\linewidth]{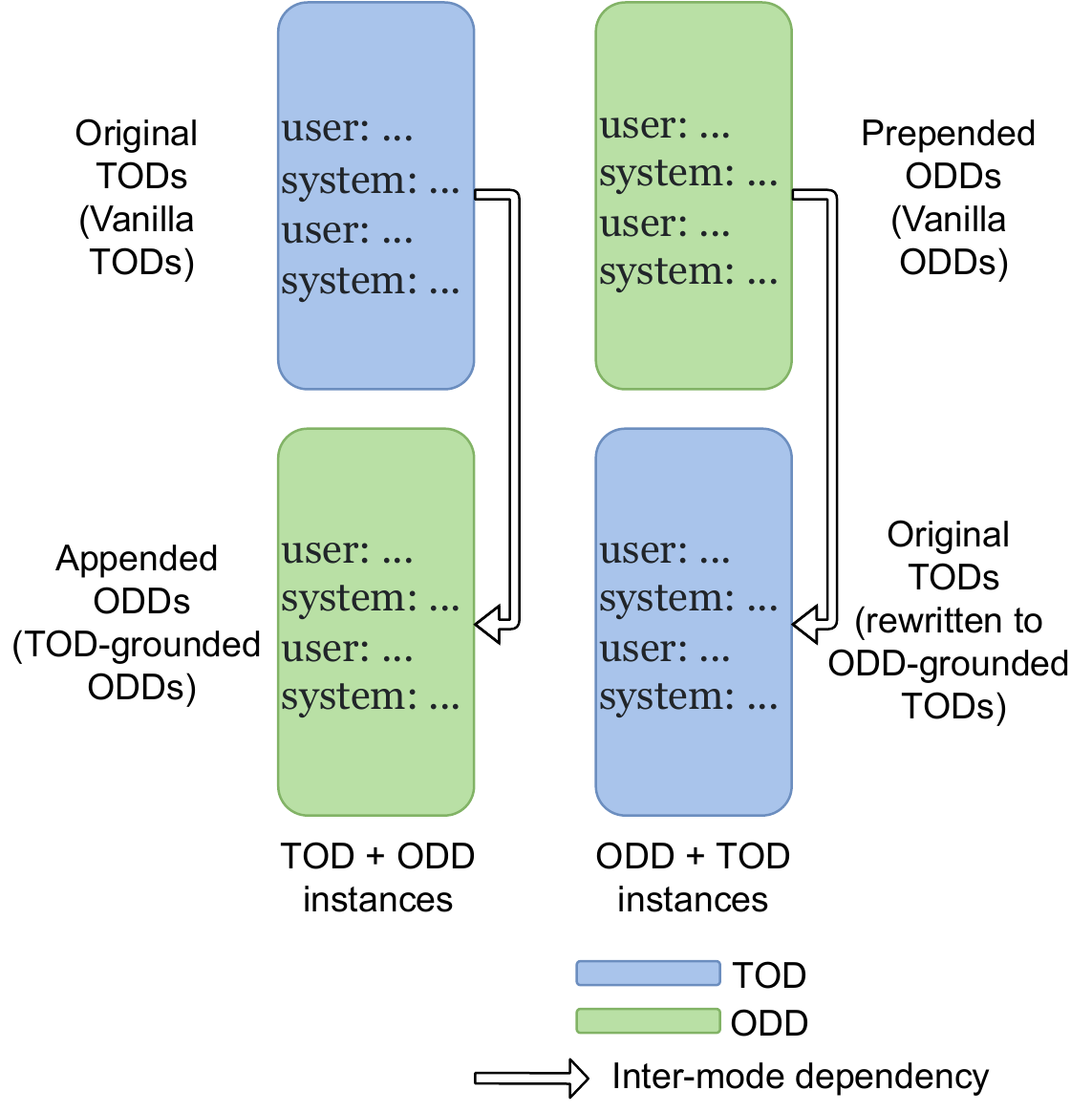}
\caption{4 types of dialogue turns are present in FusedChat, classified by the dialogue mode and the grounding context.}
\label{fig:four_types}
\end{figure}

Depending on the context and the dialogue mode, the dialogue turns in our dataset are naturally separated into 4 types in Fig.~\ref{fig:four_types}: vanilla TODs, vanilla ODDs, ODD-grounded TODs and TOD-grounded ODDs. Vanilla refers to the dialogue turns being grounded on context of its own mode only, resembling traditional datasets. The ODD turns in the ``prepending ODDs'' scenario and TOD turns in the ``appending ODDs'' scenario are vanilla.

In the following sections, we illustrate 4 unique evaluation scenarios on which FusedChat can benchmark the performance of inter-mode dialogue systems, including mode classification, TOD-grounded ODDs, ODD-grounded TODs and full inter-mode dialogues.

\subsection{Mode Classification}
The most straightforward problem one encounters in inter-mode dialogues is to decide which mode the generated response should be. Should the system respond with friendly chitchat (ODD), or should it try to interpret the user's task-oriented goal and respond with certain dialogue acts (TOD)? The accuracy for the mode classification model is shown in Table~\ref{tab:mode_class_acc}. We consider two context options: using only the latest user turn as the context (single-turn) or using the whole history containing multiple turns as the context (multi-turn). Results show that the accuracy is quite high in both cases, with ``multi-turn'' marginally outperforming ``single-turn''.

$ $

\begin{table}[h]
\centering
\begin{tabular}{|c|c|}
\hline
Context option                    & Accuracy  \\ \hline
Single-turn                     & 0.993 \\ \hline
Multi-turn                      & 0.995   \\ \hline
\end{tabular}
\caption{Mode classification accuracy for model $\mathcal{C}$.}
\label{tab:mode_class_acc}
\end{table}

\begin{table*}[h]
\footnotesize
\centering
\begin{tabular}{|c|c|c|c|c|c|}
\hline
Models               & Slot Accuracy (SA) & Joint SA & Inform & Success & BLEU  \\ \hline
\multicolumn{6}{|c|}{\textbf{ODD-grounded TODs in FusedChat}}                                            \\ \hline
\textit{Two-in-one} model          & 0.971         & 0.574               & 71.1        & 56.9         & 12.16 \\ \hline
\textit{Classification-based} model & 0.972         & 0.584               & 72.8        & 60.0         & 12.58 \\ \hline
\multicolumn{6}{|c|}{\textbf{Original MultiWOZ dataset}}                                                 \\ \hline
\textit{Neural Pipeline}~\cite{ham2020end}      & 0.976         & 0.631               & 79.2        & 64.3         & 12.72 \\ \hline
\end{tabular}
\caption{Evaluation results on ODD-grounded TODs in FusedChat and comparison with MultiWOZ results.}
\label{tab:odd-g tod}
\end{table*}

\begin{table*}[h]
\centering
\begin{tabular}{|c|c|c|c|c|}
\hline
Models & PPL & Sensibleness & Specificity  & SSA \\ \hline
\textit{Two-in-one} model  & 9.15 &     0.44         &    0.39  & 0.42 \\ \hline
\textit{Classification-based} model & 8.79 &    0.51       & 0.45  &  0.48 \\ \hline
 Ground-truth & \textit{N/A} &    0.97          &  0.91            &  0.94 \\ \hline
\end{tabular}
\caption{Evaluation results on TOD-grounded ODDs in FusedChat.}
\label{tab:tod-g odd}
\end{table*}


\begin{table*}[htbp]
\centering
    \begin{subtable}[c]{\textwidth}
    \small
        \centering
        \begin{tabular}{|c|c|c|c|c|c|}
        \hline
        Model                      & Slot Accuracy & Joint SA & Inform & Success & BLEU  \\ \hline
        \textit{Two-in-one} model           & 0.972         & 0.592    & 70.4   & 57.0    & 12.05 \\ \hline
        \textit{Classification-based} model & 0.973         & 0.600    & 75.1   & 60.9    & 12.17 \\ \hline
        \end{tabular}
        \caption{TOD Metrics \label{tbl:svmlinearaccuracy}}
    \vspace{3mm}
    \end{subtable}
  
    \begin{subtable}[c]{\textwidth}
        \centering
        \begin{tabular}{|c|c|c|c|c|}
\hline
Model                      & PPL   & Sensibleness & Specificity & SSA  \\ \hline
\textit{Two-in-one} model                & 10.49 & 0.52         & 0.47        & 0.50 \\ \hline
\textit{Classification-based} model & 10.50 & 0.58         & 0.51        & 0.55 \\ \hline
\end{tabular}
        \caption{ODD Metrics \label{tbl:svmpolynomialaccuracy}}
    \end{subtable}
 
    \caption{Evaluation results on the full FusedChat testset}
\label{tab:fusedchat}
\end{table*}

\subsection{ODD-grounded TODs}

Part of inter-mode dialogues are ODD-grounded TODs, which correspond to the ``prepending ODDs'' scenario in FusedChat. Like in regular TODs, the system's response is prompted by a task-oriented user request. However, the preceding context contains ODD exchanges, which create unique challenges. 

On the one hand, the model needs to recognize useful task-related information from the ODD context for correct dialogue state tracking. On the other hand, the system's response should correctly perform the required task-oriented function according to the latest user request, instead of derailing to chitchat by following the ODD context in the history.

Evaluation results for this portion of the dialogue turns in FusedChat are shown in Table~\ref{tab:odd-g tod}. We use the traditional TOD evaluation metrics for MultiWOZ, where slot accuracy measures dialogue state tracking, inform rate and success rate measure user goal success and BLEU measures response quality (see more details in~\cite{budzianowski2018multiwoz}).

In addition, we evaluate the \textit{Neural Pipeline} approach using the original MultiWOZ dataset (trained and tested on MultiWOZ). Remember that the \textit{classification-based} model contains $\mathcal{M}_{tod}$, which exactly follows the \textit{Neural Pipeline} model. This is to evaluate the difficulty of the new ODD-grounded TOD task compared with the vanilla TOD task in MultiWOZ. Table~\ref{tab:odd-g tod} shows that:

(1) The \textit{classification-based} model outperforms the \textit{two-in-one} model marginally.

(2) The \textit{Neural Pipeline} model evaluated on the same vanilla TOD dialogues in MultiWOZ significantly outperforms the \textit{classification-based} model evaluated on ODD-grounded TODs in FusedChat. Such significant difference suggests that ODD-grounded TODs are a more challenging task than vanilla TODs. Presumably, this is because (a) the extra requirement to correctly determine the response mode and (b) the extra need for ODD-dependent dialogue state tracking.

\subsection{TOD-grounded ODDs}

Another part of inter-mode dialogues are TOD-grounded OODs, which correspond to the ``appending ODDs'' scenario in FusedChat. The system's ODD response should be conditioned on both the TOD and ODD turns in the context.

The evaluation on ODD generation is notoriously difficult and numerous evaluation methods have been proposed~\cite{ni2021recent}. In our experiment, we follow~\cite{adiwardana2020towards} and use perplexity plus sensibleness and specificity average (SSA) as metrics. SSA represents the average between sensibleness (\textit{Does the response make sense given the context?}) and specificity (\textit{Is the response specific to the context?}). Both of them are binary for each response. A response can only be deemed specific if it is deemed sensible. SSA results are computed by averaging 5 expert human evaluators' judgement on 100 randomly sampled dialogue turns from the testset. Table~\ref{tab:tod-g odd} shows the performance of the inter-mode dialogue models on this task. 

The \textit{classification-based} model outperforms the \textit{two-in-one} model marginally. Results also show that ground-truth responses receive very high SSA scores, significantly exceeding the better dialogue model of the two we developed. This suggests that there is huge room for improvement on this task.

\subsection{Full Inter-mode Dialogues}
We show the results on the full FusedChat testset (containing all 4 types of dialogue turns) in Table~\ref{tab:fusedchat}. A combination of TOD and ODD metrics discussed above can be used to holistically gauge a dialogue system's capability to perform inter-mode dialogues. The
\textit{classification-based} model marginally outperforms the \textit{two-in-one} model.

Note that for the evaluation of ODD-grounded TODs, TOD-grounded ODDs and full inter-mode dialogues, we evaluate the response in a mode-tolerant manner. This means that even when the model generates a response of the wrong mode, we still evaluate that instance normally, instead of directly punishing the metric value to 0. For example, when evaluating BLEU, we still normally calculates the BLEU score against the ground-truth response even if the response generated by the inter-mode dialogue model is an ODD response. Of course, getting the mode wrong typically means poor scores.

\section{Related Work}

There have been multiple efforts on developing dialogue systems multi-tasking on various types of dialogues~\cite{ni2021recent}. Adapter-Bot~\cite{madotto2020adapter} uses a fixed backbone conversational model (DialoGPT) and triggers on-demand dialogue skills (e.g., empathetic responses, weather information, movie recommendation) via different adapters~\cite{houlsby2019parameter}. ~\cite{madotto2020attention} learned a dialogue system that independently parameterizes different dialogue skills, and learns to select and combine each of them through Attention over Parameters.  \citeauthor{shuster2019dialogue} (\citeyear{shuster2019dialogue}) multi-tasked on 12 separate dialogue datasets that focus on different skills and showed that a single unified model can perform decently well on all tasks. However, these works do not model the dependency between different types of dialogues in the multi-turn setting. Thus, they are not guaranteed to converse seamlessly and naturally in multiple dialogue modes simultaneously in a multi-turn conversation session.

Unlike the models trained on separate dialogue datasets,~\citeauthor{smith2020can} (\citeyear{smith2020can}) tried to fuse multiple skills into one conversation session. They built a new dialogue dataset named Blendedskilltalk containing dialogues where knowledge, emotional and personalizing skills are shown together in the same multi-turn conversation. They showed that systems fine-tuned on the new multi-skill dataset have improved ability in handling multiple skills simultaneously in the same multi-turn conversation session. However, they only targeted open-domain conversations. Our work, on the other hand, targets the fusion of general ODDs and TODs, as we view them as the two most mainstream forms of dialogues for the research community currently. Along the direction of fusing TODs and ODDs,~\citeauthor{zhao2017generative} (\citeyear{zhao2017generative}) proposed to artificially augment TODs with randomly sampled utterances from a chitchat corpus, mainly to improve the out-of-domain recovery performance for the TOD system.

\citeauthor{sun2020adding} (\citeyear{sun2020adding}) proposed to decorate TOD responses with ODD snippets, in order to make the dialogue agent sound more engaging and interactive. Unlike~\cite{sun2020adding}, where ODD snippets act as a supplementary role to TOD responses, our dataset tackles the fusion of TODs and ODDs by treating them as parallel dialogue modes of equal importance, and focuses on modeling inter-mode dependency in the multi-turn setting.

There were efforts that followed our work that further explored dialogue systems with multiple utilities. \cite{chen2022ketod} additionally conditioned the chit-chat to knowledge snippets. Dialogue systems under their consideration are capable of TODs, chit-chat based on commonsense, and additionally chit-chat based on knowledge snippets. \cite{li2022opera} created a new dataset OB-MultiWOZ, where TOD sessions are enriched with QA-like information seeking grounded on external knowledge.

\section{Chapter Summary and Future Prospects}

Our work demonstrated in this chapter serves the goal to develop dialogue systems that are capable of performing both TODs and ODDs with inter-mode dependency\footnote{The work in this chapter has been published in \cite{young2021fusing}}. Compared with traditional datasets, the new dataset FusedChat uniquely contains ODD-grounded TODs and TOD-grounded ODDs. It endeavors to fuse the two common forms of human conversations, i.e., casual open-ended conversations supported only by commonsense, and task-oriented conversations supported by specific knowledge bases. We show preliminary experiment results on two baseline models, which suggest huge room for improvement. We release dataset and baselines in order to propel future work on inter-mode dialogue systems.

We note that the framework set by FusedChat is limited. The dataset does not contain dialogue sessions containing more than one mode switch, which represents a gap with real-world scenarios. We suspect more mode switches could make inter-mode dialogues even more challenging. Our choice of TODs and ODDs does not represent the full scope of possible dialogue settings. We chose the most simple form of ODDs where the response is only determined by the context. Yet in the literature, ODDs have been grounded on various forms of information, such as personas~\cite{zhang2018personalizing}. We chose the classical setting of TODs as in MultiWOZ, which is defined by structured entity-centric knowledge bases. However, the concept of TODs has seen expansion, such as with unstructured knowledge access~\cite{kim2020beyond}. We expect the fusion of more complex forms of ODDs and TODs to be more challenging, but they would even better represent human-level conversational abilities.

The construction of FusedChat required a lot of manual creative effort. It is thus very expensive to replicate the same routine for every new inter-mode dialogue scenario. Alternatively, zero-shot or few-shot models that can learn to perform inter-mode dialogues by mostly relying on separate single-mode dialogues are a promising direction. FusedChat can also serve as a test-bed for such paradigms.

\chapter{Grounding dialogues on the audio modality} 
\chaptermark{Grounding dialogues on the audio modality}  
\label{ch:audio-chapter}

Effort on developing dialogue systems started out assuming text-based conversation, where the user message was modeled as a sequence of words in a vocabulary. Real-world human conversation, in contrast, involves other modalities, such as voice, facial expression and body language, which can influence the conversation significantly in certain scenarios. This chapter demonstrates our effort on exploring the impact of incorporating the audio features of the context into generative dialogue systems. Specifically, we first design an auxiliary response retrieval task for audio representation learning. Then we use word-level modality fusion to incorporate the audio features as additional context to our main generative model. Experiments show that our audio-augmented model outperforms the audio-free counterpart on perplexity, response diversity and human evaluation.

\section{Introduction}


There are many ways that audio signals play a role in conversation. Audio signals naturally carry emotional information. For example, ``Oh, my god!'' generally expresses surprise. But depending on the voice shade, a wide range of different emotions can also be carried, including fear, anger and happiness. Audio signals can have strong semantic functions as well. They may augment or alter the meaning expressed in text. For example, ``Oh, that's great!'' usually shows positive attitude. But with a particular voice shade of contempt, the same utterance can be construed as sarcastic. Stress also plays a role in semantics: ``I think \textit{she} stole your money'' emphasizes the speaker's opinion on the identity of the thief while ``I think she stole \textit{your} money'' emphasizes the speaker's opinion on the identity of the victim.

Therefore, while identical from a written point of view, utterances may acquire different meanings based solely on audio information. Empowering a dialogue system with such information is necessary to interpret an utterance correctly and generate an appropriate response.

In this chapter, we explore dialogue generation augmented by the audio modality under the commonly-used Seq2Seq framework. First, because of the noisiness of the audio signal and the high dimensionality of raw audio features, we design an auxiliary response classification task to learn suitable audio representation for our dialogue generation objective. Second, we use word-level modality fusion for integrating audio features into the Seq2Seq framework. We design experiments to test how well our model can generate appropriate responses corresponding to the emotion and emphasis expressed in the audio. They show that our model captures the following phenomena in conversation: Vocally emphasized words in an utterance are relatively important to response generation; and emotion expressed in the audio of an utterance has influence on the response.

\section{Proposed Approaches}

\subsection{Audio Representation Learning}\label{subsec-audioreplearn}

Raw features extracted from audio sequences are high-dimensional and noisy. They are not suited as direct input to the dialogue generative model. For example, the number of dimensions for word embeddings used in RNNs is typically below 648. However, the number of raw audio features can reach 10000 \cite{schuller2013interspeech}. 

Therefore, we need an audio representation learning method to reduce the number of dimensions of the audio features and also make them suitable for the dialogue generation task.

For this purpose, we design an auxiliary response classification task based on audio features.

Specifically, we construct a set of ${<}context, response, label{>}$ triples, where $label$ is binary indicating whether the context and response combination comes from a real conversation dataset $D$ or is randomly assembled as a negative example. The goal of this task is to predict $label$ based on the ${<}context, response{>}$ pair.

Following \cite{lowe2015ubuntu}, our classification model is defined as:
\begin{eqnarray}
\label{eq:compatibility}
f(x,y) = sigmoid(\bm{x}^{T}\bm{W}\bm{y}),
\end{eqnarray}
      where $\bm{x}$ and $\bm{y}$ are representations of the context $x$ \footnote{The meaning of a mathematical symbol stays the same in the same chapter. In different chapters we may use the same symbol for different meanings in order to keep the symbol set simple.} and response $y$ respectively. Matrix $\bm{W}$ is the model parameter.

\begin{figure*}
    \centering
    \includegraphics[width=13cm]{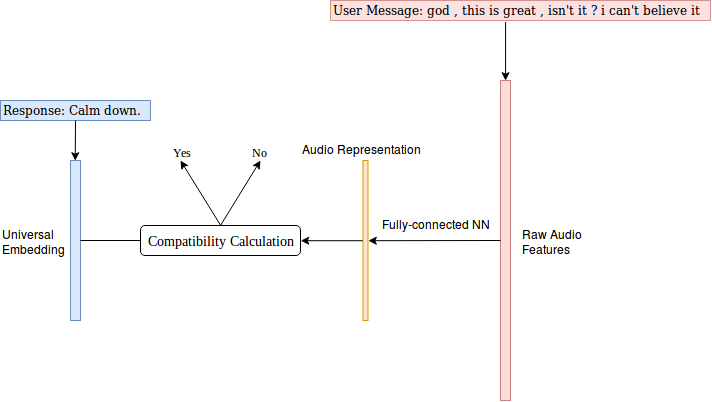}
    \caption{A response classification model is
    used as the auxiliary task for audio representation learning.}\label{fig:response_selection}
\end{figure*}

We use a universal sentence encoder \cite{conneau2017supervised} for the representation of response $\bm{y}$. For the purpose of finding the best audio context representation, $\bm{x}$ is determined only by audio features $\bm{a}_i$ of individual words in the context:
\begin{eqnarray}
\bm{c} = avg(P(\bm{a}_i)), i\in[0, n),
\end{eqnarray}

where $P$ is a perceptron and $n$ is the number of words in the context. The model is shown in Figure \ref{fig:response_selection}.

This model is trained on a conversation dataset $D$ for best classification accuracy using mean squared loss between $label$ and $f(x,y)$ in Equation (\ref{eq:compatibility}). After training, the output of the perceptron $\tilde{\bm{a}}_i = P(\bm{a}_i)$ is taken as the word-level audio representation used in the generative dialogue systems.

Note that separately learning audio representations also serves the purpose of reducing memory burdens when the main generation model is trained.

\subsection{Audio-augmented Seq2Seq Model}

\begin{figure*}
    \centering\includegraphics[width=11cm]{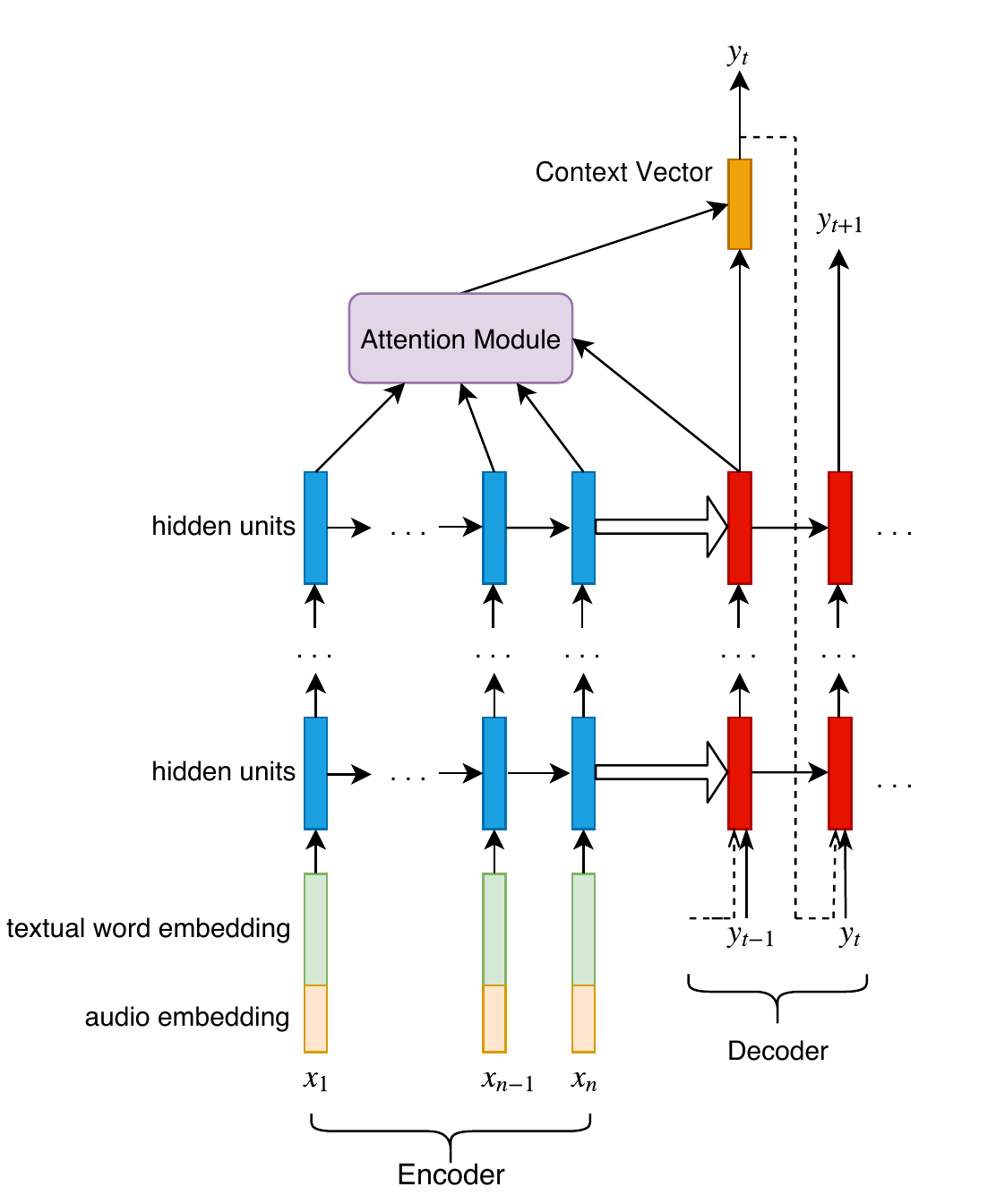}
    \caption{Audio-Seq2Seq model}\label{fig:audio-seq2seq}
\end{figure*}

We build upon the general encoder-decoder framework which is based on sequence to sequence learning \cite{sutskever2014sequence}. The encoder represents a user message (context) $x = x_1x_2\cdots x_{n}$ with hidden representations $\bm{H} = \bm{h}_1\bm{h}_2\cdots \bm{h}_n$, which is briefly defined as below:

\begin{equation}
\bm{h}_n = \mathbf{LSTM_E}(\bm{h}_{n-1},\bm{e}(x_n)),
\end{equation}

where $\mathbf{E}$ denotes encoder. The decoder takes as input a context vector $\bm{\bar{c}}_{t-1}$ produced by an attention mechanism and the embedding of a previously decoded word $\bm{e}(y_{t-1})$, and updates its state $\bm{s}_t$ using another LSTM: 
\begin{equation}
\bm{s}_t = \mathbf{LSTM_D}(\bm{s}_{t-1},[\bm{\bar{c}}_{t-1};\bm{e}(y_{t-1})]),
\label{eq:decodersimple}
\end{equation}
where $\mathbf{D}$ denotes decoder. The decoder generates a token by sampling from the output probability distribution which is determined by $\bm{\bar{c}}_{t}$.
%

Following \cite{chen2017multimodal}, we use a simple word-level embedding concatenation method for integrating audio features into word representation:
\begin{equation}
\bm{e}(x_n) = [\bm{w}_n; \tilde{\bm{a}}_n], \end{equation}
where $\bm{w}_n$ is the traditional word embedding and $\tilde{\bm{a}}_n$ is the word-level audio representation. Thus, in our new Audio-Seq2Seq model (Figure \ref{fig:audio-seq2seq}), the word representation contains both textual and audio information. 

\section{Experiments and Results}

\subsection{Dataset}
Most of the existing and consolidated datasets used in dialogue system related research come with textual content only \cite{lowe2015ubuntu,ritter2010unsupervised}. Fortunately, along with the growing interest in multimodal systems, some datasets are fit for our task. We experiment with two such datasets, the Interactive Emotional Dyadic Motion Capture dataset (IEMOCAP) \cite{busso2008iemocap} and the Multimodal EmotionLines dataset (MELD) \cite{poria2018meld}.

IEMOCAP was designed with the main intent of providing a corpus of dyadic conversations capable of conveying emotions. Two types of dialogue sessions were created for IEMOCAP to achieve this task: scripted and spontaneous sessions. In the scripted case, two actors, a male and a female, were asked to rehearse some previously memorized scripts, as this supposedly leads to a more genuine expression of emotions than directly reading off a script. 
In the spontaneous case, the actors were given more liberty to use their own words to discuss about selected emotion-evoking topics. This supposedly allows the actors to express more natural emotions. 
The dataset contains a total of 10039 utterances with their corresponding audio segments.

MELD is a dataset containing utterances from the TV series \textit{Friends}. For each utterance multimodal information in the form of text, audio and video is provided. MELD consists of 1433 dialogues for a total of 13708 utterances.

These two datasets are suitable for our purposes as the audio component is strongly representative of the speaker's emotional state and plays a pivotal role in the meaning to be conveyed.

\subsection{Experiment Details}

\subsubsection{Data Preprocessing}

From IEMOCAP and MELD set of dialogues we extract \textit{\textless sentence, response\textgreater} pairs by taking successive utterances within individual dialogues. Formally, from dialogue $d_i = \{u_1,..,u_{n_i}\}$, where $u_1,..,u_{n_i}$ are the utterances composing the dialogue, we extract the set of pairs $\{{<}u_1, u_2{>}, .., {<}u_{{n_i}-1}, u_{n_i}{>}\}$.
From the resulting pairs we create a vocabulary, for each of the datasets, containing only the terms with more than one occurrence in the respective corpus and that are present in the standard English vocabulary \cite{pyenchant} and those that are not present in the English vocabulary but occur ten or more times in the dataset. Our final vocabulary sizes are $2171$ for IEMOCAP and $3123$ for MELD. After these procedures we end up with a total of $7901$ utterances for IEMOCAP and $12274$ for MELD.
\begin{table}[]
\centering
\begin{tabular}{|l|l|l|}
\hline
\textbf{}                & \textit{IEMOCAP} & \textit{MELD}  \\ \hline
\textit{No. Utt.}        & 7901             & 12274          \\ \hline
\textit{Avg Utt. Length} & 15.26            & 10.69 \\ \hline
\textit{Train. set size}    & 6000             & 10000           \\ \hline
\textit{Dev. set size}    & 1000             & 1174           \\ \hline
\textit{Test set size}    & 901             & 1000           \\ \hline
\textit{Vocabulary size} & 2171             & 3123           \\ \hline
\end{tabular}
\caption{Number of utterances, average length of utterances, development set sizes, test set sizes and vocabulary sizes for IEMOCAP and MELD datasets.}\label{tab:datasets}
\end{table}

The audio segments provided within the datasets are given at a sentence granularity. We therefore conduct word alignment and obtain word-level audio features. We first use the GENTLE forced aligner \cite{ochshorn2016gentle} to find the start and end timestamps of each word within a sentence. Then, with OpenSMILE \cite{eyben2010opensmile}, we extract 6373 raw audio features for each word. We use the \textit{IS13\_ComParE.conf} configuration \cite{schuller2013interspeech} that has been widely used in emotion recognition tasks \cite{poria2018meld, xianyu2016svr}, rendering it a suitable choice for our case, as impacting the conveyed emotion is one of the primary ways audio features make an impact in conversation.
We randomly sample utterances from the datasets to split into training, development, and test sets. In Table \ref{tab:datasets} we report some of the most important statistics regarding the datasets we operate on.

\subsubsection{Model training details}

In our audio representation learning model (Section \ref{subsec-audioreplearn}), the response sentence embedding given by the universal sentence encoder has size $4096$. During the training process, the best audio representation extractor is obtained at the point when the classification accuracy on the development dataset is the highest.

We use the Seq2Seq model with Luong attention mechanism \cite{luong2015effective} as the backbone of our main audio-augmented model. It is a pruned version of the main model (Figure \ref{fig:audio-seq2seq}) that does not use audio features in its word-level representation. After being trained on a large text-based conversation dataset, the resulting model parameters are transferred to the main model as initialization of its parameters corresponding to textual input.

A 3.3M Reddit Conversation Dataset \cite{reddit} is used for this purpose. We filter it using the vocabularies previously created for the audio conversation datasets. Specifically a conversation pair ${<}u_j, u_{j+1}{>}$ is removed if it contains more than one out-of-vocabulary term.

We follow \cite{luong17} for most of the hyperparameter settings. All generation models are trained for 50000 steps with batch size 256 after initialization with the pretrained model. The learning rate is set to 0.1. Dropout rate is 0.3. We use 2 layers of hidden units. Keeping audio representation fixed, we search for the optimal text embedding dimension in ${10, 25, 50, 100}$. 100 yields the best results.
By manually inspecting the generated responses at different steps, we find that they are most natural-sounding when the models slightly overfit. In contrast, the models generate overly simple responses when the development perplexity is lowest. This might be due to the fact that the audio conversation datasets are relatively small. We manually choose the best checkpoint for testing based on human perception of response quality on the development set after the models start overfitting.

\subsection{Experiment Results}

\subsubsection{Results on audio representation learning} \label{sec:arl}

The results are shown in Table \ref{tab:nneres}. The fact that the accuracies are much higher than 50\% indicates that audio features indeed carry information that is relevant to conversation.
Overall, the accuracies show only a slight improvement in spite of a substantial increase in dimension moving up from $25$. We choose $25$ as the size of the audio representation that is adopted in all the experiments.

\begin{table}
\centering
\begin{tabular}{|l|l|l|l|}
\hline
\backslashbox{\textbf{Dataset}}{\textbf{Dimension}} & \textit{25} & \textit{50} & \textit{100} \\ \hline
\textit{IEMOCAP}  & 59.4\%      & 62.4\%      & 61.8\%       \\ \hline
\textit{MELD}     & 54.8\%      & 54.8\%     & 54.6\% \\ \hline
\end{tabular}
\caption[Response classification accuracy]{Auxiliary response classification task accuracy varying the dimension of the audio representation.}\label{tab:nneres}
\end{table}

\subsubsection{Perplexity, Diversity and Human Evaluation}


To  In this work we consider two automatic metrics in addition to human judgment: perplexity and diversity. Lower perplexity on the testset indicates that the model fits the testset better. \textit{Diversity} is defined as the number of unique words generated by the model over the test set. Lack of diversity and tendency to generate similar, short responses regardless of the different inputs is a notorious problem in generative conversational models \cite{liu2018knowledge}. A model that generates interesting and information-rich responses is characterized by high diversity. 
Automatic and human evaluation results are shown in Tables \ref{statIEMO} and \ref{statMELD}.  ``$\pm$ value'' indicates standard deviation. We see that the Audio-Seq2Seq model achieves lower perplexity and higher diversity on both datasets. Since adding audio features essentially enriches the representation of input to the Seq2Seq model, it helps the model generate the correct output (lower perplexity). Also, the additional audio information increases the diversity of the input, which helps generate diverse responses (higher diversity).

\begin{table}
\centering
\begin{tabular}{|c|c|c|c|}
\hline
\backslashbox{\textbf{Model}}{\textbf{Metric}}        & \multicolumn{1}{c|}{\textit{Perplexity}} & \multicolumn{1}{c|}{\textit{Diversity}} & \multicolumn{1}{c|}{\textit{Human Preference}} \\ \hline
\textit{Seq2Seq}      &     36.83 $\pm$ 0.34      &    805 $\pm$ 10.5       & 44.4\%                                \\ \hline
\textit{Audio-Seq2Seq} &     31.13 $\pm$ 0.31       &   831 $\pm$ 12.8        & 55.6\%                                \\ \hline
\end{tabular}
\caption{Statistics on IEMOCAP.}\label{statIEMO}
\end{table}

\begin{table}
\centering
\begin{tabular}{|c|c|c|c|}
\hline
\backslashbox{\textbf{Model}}{\textbf{Metric}}        & \textit{Perplexity} & \textit{Diversity} & \multicolumn{1}{l|}{\textit{Human Preference}} \\ \hline
\textit{Seq2Seq}      &     47.83 $\pm$ 0.44       &   567 $\pm$ 8.7        & 46.5\%                             \\ \hline
\textit{Audio-Seq2Seq} &      46.19 $\pm$ 0.49      &   629 $\pm$ 10.0        & 53.5\%                                \\ \hline
\end{tabular}
\caption{Statistics on MELD}\label{statMELD}
\end{table}

\begin{table}
\centering
\begin{tabular}{|c|c|c|}
\hline
\textbf{}                 & \textit{Audio-Seq2Seq} & \textit{Seq2Seq} \\ \hline
\textit{IEMOCAP}          & 354                    & 283     \\ \hline
\textit{MELD}             & 359                    & 313              \\ \hline
\end{tabular}
\caption[]{Number of responses by the corresponding model that are deemed as better according to AMT judges. See the correponding percentages in Tables \ref{statIEMO} and \ref{statMELD}.}\label{tab:ch4-avsnoa}
\end{table}

A number of samples were manually selected from the test set for human evaluation. During this process, we only chose samples with high-quality user messages, considering both audio and text. A sample is excluded if the audio is noisy or the user message is too short or generic.

We conducted pair-wise comparisons between responses generated by the Audio-Seq2Seq model and the Seq2Seq model. Seven judges were recruited on the Amazon Mechanical Turk (AMT) platform. After listening to the audio of the user message, the judges were asked to select the best response from those generated by the two models.
We asked the judges to follow two rules: (1) Prefer the response that is compatible with the tone of the user message. (2) Prefer longer and more informative responses. The results of the judges' evaluation are shown in Table \ref{tab:ch4-avsnoa}.

Table \ref{tab:cases} shows cases where the responses produced by the Audio-Seq2Seq model suit the audio context better. For example, in the second case, the voice of the user message shows excitement and restlessness, which is captured by the ``calm down'' in the response.

\begin{table}[]
\centering
\scriptsize
\begin{tabular}{|c|c|c|}
\hline
Context                                                                                                                                                                                             & Seq2Seq Response & Audio-Seq2Seq Response                                                                                                                \\ \hline
\begin{tabular}[c]{@{}c@{}}because i know you’re at a desk right now \\ in your corporate building and \\ you’re going to send me a check for sixteen \\ dollars and forty two cents!\end{tabular} & right, ok.      & \begin{tabular}[c]{@{}c@{}}oh, no, no, no, no, no. \\ i don’t know. i don’t know. \\ it’s not what i can do.\end{tabular} \\ \hline
\begin{tabular}[c]{@{}c@{}}what time is it? god, this is great, \\ isn’t it? i can’t believe it.\end{tabular}                                                                                  & yeah, it is      & okay, all right. just just calm down                                                                                                \\ \hline
\begin{tabular}[c]{@{}c@{}}no. it’s his newspapers is what it is now \\ every month there’s a new boy comes home \\ and larry’s going to be the next one.\end{tabular}                            & what?           & alright. alright. alright just listen.                                                                                             \\ \hline
\end{tabular}
\caption{Samples produced by Audio-Seq2Seq. Considering the tone of the context, the responses Audio-Seq2Seq produced are more appropriate.}\label{tab:cases}
\end{table}

\subsubsection{Emotional appropriateness}

Audio features can be strongly indicative of the speaker's emotion, and thus have influence on the response.
In order to quantitatively evaluate the ability of our model to generate emotionally appropriate responses, we design an experiment with an artificially constructed set of 200 audio message samples of two different emotional states. Specifically, 100 textual message samples are selected from the test set and for each sample we manufacture two audio segments of different emotions by availing ourselves of the MARY Text-To-Speech system (MaryTTS) \cite{Schroeder2001}.

Following Russell's Circumplex model of affect \cite{russell1980circumplex} we vary the valence dimensions of the synthesized audio segments. With arousal and valence in the range $[0,1]$, we use a fixed arousal value of $0.9$ combined with the two valence values $0.1$ and $0.9$. When valence = $0.9$, the synthesized speech is fast and highly-pitched, exhibiting an excited emotional state. Whereas when valence = $0.1$, the synthesized speech is slow and calm.
Our Audio-Seq2Seq model generates two responses corresponding to those two audio segments of different emotion states. To evaluate how well a response matches an emotional state, we shuffle the two responses and ask human judges to match audio segments with the responses to see if the results agree with the model's.

This association task performed by three judges shows that human evaluation tends to agree with the responses generated by the model more often than random guess. Details are given in Table \ref{tab:marytable}. Table \ref{tab:marytablesamples} shows cases where the model seems to be able to perceive the emotional state of the speaker and adapt its response accordingly. When the audio expresses an excited state (valence = $0.9$), the model is able to tune its response in a suitable manner. For instance in the first sample the second response shows a strong correlation with the excited and agitated state of the speaker by asking him to calm down. In the second sample, the higher rate with which the valence = $0.9$ context is uttered due to the excited state makes the speaker sound less sincere thus eliciting a stuttered and complaining response as compared to the more composed and calm one when valence = $0.1$.

\begin{table}[h]
\centering
\begin{tabular}{|c|c|c|c|}
\hline
\textbf{Model}       & \textit{Agree} & \textit{Disagree} & \multicolumn{1}{l|}{\textit{Cannot determine}}\\ \hline
\textit{IEMOCAP}     & 25.4\%           & 15.1\%        & 59.5\%                \\ \hline
\textit{MELD}        & 28.2\%           & 16.7\%        & 55.1\%                 \\ \hline
\end{tabular}
\caption{Percentage of cases on which the judges' verdicts agree or disagree with the model. The number of agreement cases exceeds the number of disagreement cases. This indicates that to a certain degree, the model's response captures the emotion expressed in the audio features of the message.}\label{tab:marytable}
\end{table}

\begin{table}[h]
\centering
\begin{tabular}{|c|c|c|}
\hline
\textit{Context}                                                                          & \textit{Valence = 0.1} & \textit{Valence = 0.9}                                                                                                        \\ \hline
\begin{tabular}[c]{@{}c@{}}turn it off \\ it's driving me mad.\end{tabular} &  I won't.      & \begin{tabular}[c]{@{}c@{}}well, do try to \\control yourself darling.\end{tabular} \\ \hline
\begin{tabular}[c]{@{}c@{}}okay that's helpful.\\ thanks. \end{tabular}                                                                                  & \begin{tabular}[c]{@{}c@{}}i've been trying to \\work this backwards   \end{tabular}   & this is all this is unfair.                                                                                            \\ \hline
\end{tabular}
\caption{The model adapts its response based on the emotion expressed in the message through audio features.}\label{tab:marytablesamples}
\end{table}

\subsubsection{Attention on vocally emphasized words}

In a conversation, vocally emphasized words in an utterance are most important to information communication. To evaluate how well our model captures this phenomenon, we calculate the correlation between the volume/duration of the audio segments of words in the user message and the attention the words get during the generation process.

We take the length of the audio segment of an individual word as the \textit{duration} of that word and \textit{maximum amplitude} is used to indicate \textit{volume}.

For calculating attention on a word in the message, we sum all attention scores it gets during the response generation process. Specifically, for the generation of response word $y_t$, the attention score on message word $x_i$ is $a_{it}$. For the generated response $[y_1, y_2, ... ,y_n]$, the total attention on $x_i$ is $a_{i} = \sum_{t=1}^{n} a_{it}$.

We normalize attention, duration and maximum amplitude by dividing them by average values over the message. Pearson and Spearman correlations are calculated on attention-duration and attention-maximum amplitude pairs. The results are shown in Tables \ref{tab:correlation1} and \ref{tab:correlation2}. On both datasets our experiment shows relatively strong positive correlation between attention and duration. For attention and maximum amplitude, however, our calculation only shows slightly positive correlation. This implies that in our dataset, length is more indicative of a word’s importance to the dialogue system than volume. But this observation cannot be generalized without more experiments on more datasets.

\begin{figure}
    \centering
    \includegraphics[width=10cm]{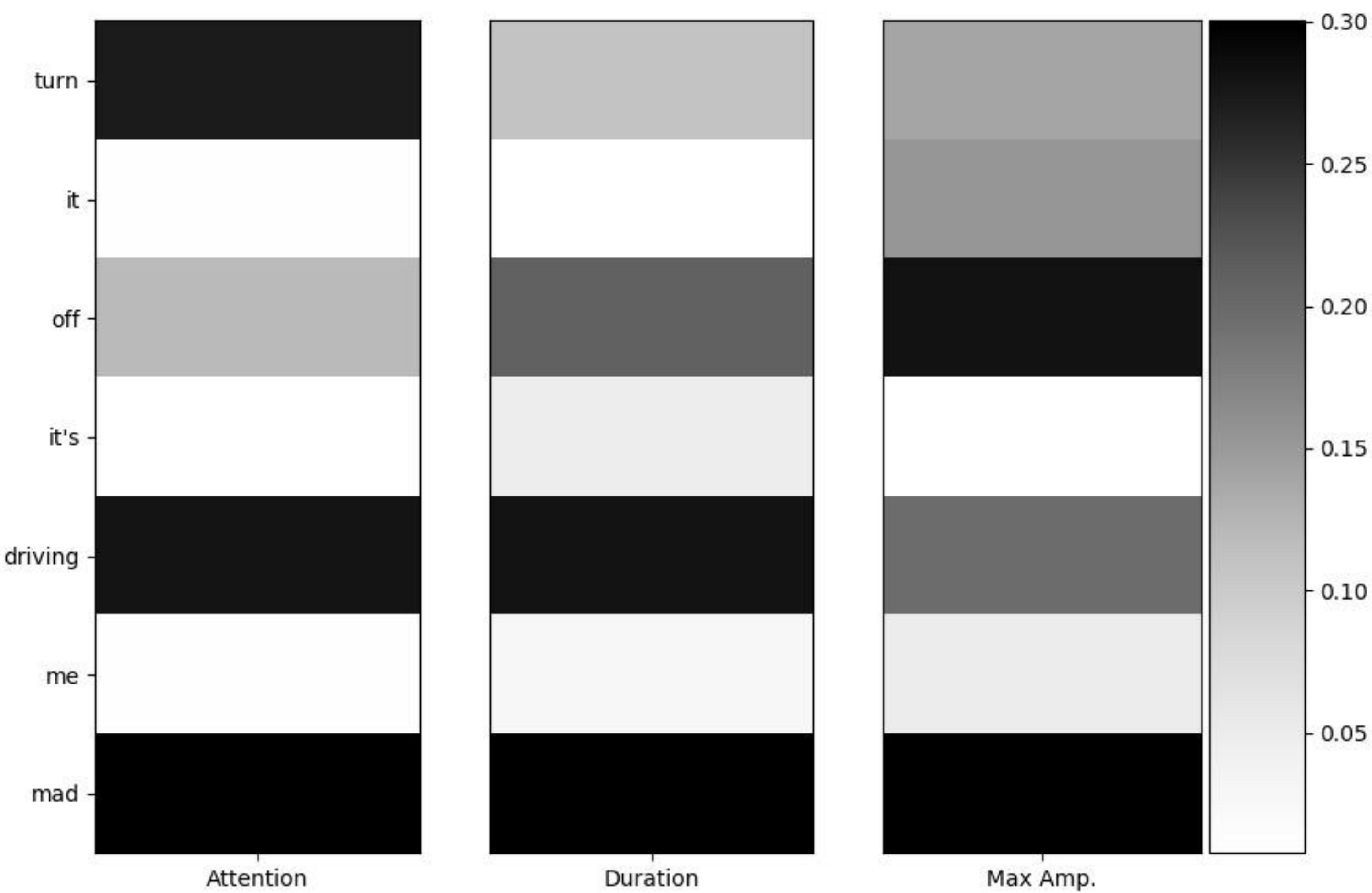}
    \includegraphics[width=10cm]{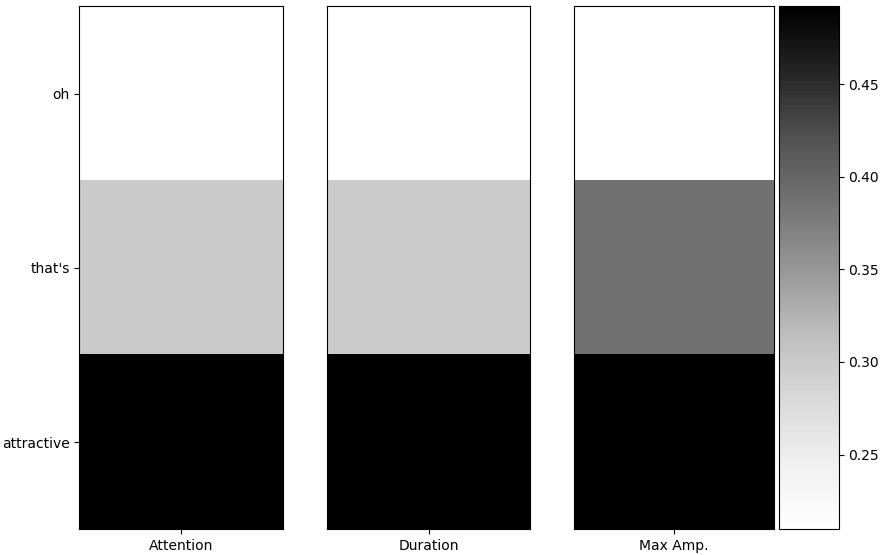}
    \caption{The attention of a word in the source sequence is positively correlated with both of its duration and maximum amplitude.}\label{attention}
\end{figure}

Two examples are shown in Figure \ref{attention}. 
In the message ``turn it off it's driving me mad'', ``off'', ``driving'' and ``mad'' are vocally emphasized. Accordingly, attention scores on those three words are relatively high. In a shorter example, ``oh that's attractive'', the word ``attractive'' contains the most semantic information. It is vocally emphasized and gets the most attention.

\begin{table}[h]
\centering
\begin{tabular}{|l|l|l|l|l|}
\hline
\textbf{IEMOCAP}            & \textit{Pearson's r}   & \textit{Spearman's }$\rho$  \\ \hline
\textit{Attention/Duration} & 0.418                                         & 0.384 \\ \hline
\textit{Attention/Max Amp.} & 0.096                 & 0.128                     \\ \hline
\end{tabular}
\caption[MELD attention/word stress correlation]{The correlation between word attention and duration/maximum amplitude on IEMOCAP.}\label{tab:correlation1}
\end{table}
\quad
\begin{table}[h]
\centering
\begin{tabular}{|l|l|l|l|l|}
\hline
\textbf{MELD}            & \textit{Pearson's r} & \textit{Spearman's }$\rho$  \\ \hline
\textit{Attention/Duration} & 0.312                & 0.334                   \\ \hline
\textit{Attention/Max Amp.} & 0.094                & 0.069                   \\ \hline
\end{tabular}
\caption[MELD attention/word stress correlation]{The correlation between word attention and duration/maximum amplitude on MELD.}\label{tab:correlation2}
\end{table}

\section{Related Work}

As a research effort on developing dialogue systems that are additionally conditioned on the audio modality in 2019, our work was inspired by previous work on multimodal NLP and representation learning. At the same time, we gladly notice following work that further explores multi-modal dialogue systems from 2020 to 2022. This section discusses both categories of related work.

{\large \textbf{Preceding work}}



Human conversation naturally involves multiple modalities. This important fact had been noticed in research preceding our work demonstrated in this chapter. 

First, the subject/background of a conversation can be multimodal. For example, in image-grounded conversation \cite{mostafazadeh2017image}, two interlocutors generate conversations based on a shared image. For this task, visual features of the image need to be infused into the context vector. \cite{alamri2018audio} proposed Visual Scene-aware Dialogs, a scenario where the dialogue system discusses dynamic scenes with humans. A scene, in the form of a short video, is presented to the interlocutors as the conversational context. For this task, \cite{hori2018end} incorporated techniques for multimodal attention-based video description into an end-to-end dialogue system. Audio and visual features that come from deep video description models are used to augment the context vector. \cite{saha2018towards} proposed a large domain-aware multimodal conversation dataset where shoppers and sales agents converse about products in the fashion domain. Each conversational turn is composed of text and corresponding images being referred to. For this scenario, \cite{agarwal2018improving} proposed a multimodal extension to the Hierarchical Recurrent Encoder-Decoder (HRED) \cite{serban2016building} for in-turn multimodality and multi-turn context representation.

Second, human conversation itself involves multiple channels of information. Voice, body language and facial expressions all play roles in conversation. In an ideal human-machine conversational system, machines should understand this multimodal language. This information had seen use in conversation analysis. \cite{yu2015attention} proposed to model user engagement and attention in real time by leveraging multimodal human behaviors, such as smiles and speech volume.
\cite{gu2018human} performed emotion recognition, sentiment analysis, and speaker trait analysis on conversation data using a hierarchical encoder that formulates word-level features from video, audio, and text data into conversation-level features with modality attention.

Our method of word-level modality fusion had already seen use in multimodal sentiment analysis. In \cite{chen2017multimodal}, the RNN, which acts as the utterance encoder, takes a concatenation of audio, video and text features as input at every time step. On the Interactive Emotional Dyadic Motion Capture (IEMOCAP) dataset \cite{busso2008iemocap}, \cite{chen2017multimodal} showed considerable improvement on dialogue emotion classification accuracy by integrating audio features. This result motivated our work - since incorporating audio features improves emotion classification accuracy in conversation and emotion is useful to response generation \cite{zhou2018emotional}, we hypothesized that incorporating audio features improves response generation.

{\large \textbf{Following work}}

We would also like to note works done from 2020 to 2022 following our work along the research line of multi-modal dialogue systems. 

First, following the recent breakthrough on large-scale self-supervised pretraining for language models \cite{devlin2018bert, brown2020language}, multi-modal learning has adopted the same routine. \cite{akbari2021vatt} presented a framework for learning multimodal representations from unlabeled data using Transformers. Specifically, their Video-Audio-Text Transformer (VATT) takes raw signals as inputs and extracts multimodal representations that benefit various downstream tasks. They trained VATT end-to-end from scratch using multimodal contrastive losses. Its usefulness was proven based on video action recognition, audio event classification, and text-to-video retrieval. Data2vec \cite{baevski2022data2vec} was another attempt at performing self-supervised learning across modalities. It uses the same learning method for either speech, NLP or computer vision. It predicts latent representations of the full input data based on a masked view of the input in a self-distillation setup using a standard Transformer architecture.  Instead of predicting modality-specific targets such as words, visual tokens or units of human speech which are local in nature, data2vec predicts contextualized latent representations that contain information from the entire input.

Second, researchers have come up new scenarios that further enrich multi-modal dialogues. For example, \cite{kamezawa2020visually} proposed a visually-grounded first-person dialogue dataset with verbal and non-verbal responses. It provides manually annotated first-person images and eye-gaze locations of the speakers. \cite{moon2020situated} envisioned dialogue systems to both take multimodal inputs and perform multimodal actions. Situated Interactive Multi-Modal Conversations (SIMMC) was introduced as a new dataset aimed at training agents that take multimodal actions grounded in a co-evolving multimodal input context in addition to the dialogue history. \cite{yang2021unimf} proposed a new dataset that conditions end-to-end task-oriented dialogue systems on multimodal knowledge bases.

Third, new approaches have been proposed based on the two advances above. \cite{le2020video} leveraged the power of pre-trained language models for improving video-grounded dialogues. GPT-2 was extended to tackle these challenges by formulating the video-grounded dialogue task as a Seq2Seq learning task, combining both visual and textual representation into a structured sequence. \cite{li2021bridging} proposed a very similar approach with GPT-2. Their difference lies in their choices of fine-tuning tasks. They both fine-tune the model for the traditional response generation task. Additionally, \cite{le2020video} fine-tuned their model for masked multi-modal modeling, and to match video-text pairs. \cite{li2021bridging} on the other hand, chose video-audio feature regression and caption generation.

\section{Chapter Summary and Future Prospects}

In this chapter, we augment the common Seq2Seq dialogue model with audio features and show that the resulting model outperforms the audio-free baseline on several evaluation metrics. It also captures interesting audio-related conversation phenomena\footnote{The work in this chapter has been published in \cite{young2020dialogue}.}.

Although only using text in dialogue systems is a good enough approximation in a lot of scenarios, other modalities have to be integrated before automatic dialogue systems can reach human performance. Our work belongs to such a line of research that strives to build multimodal dialogue systems.

We view multimodal dialogue systems as a very promising field worthy of future investigation. We believe it implies immediate application value and at the same time strongly relates to the long-term success of the field of AI in general. One future direction that we believe may bear immediate fruit is to utilize large-scale multimodal pretraining on conversational data to improve multimodal dialogue systems. For example, one might consider explicitly injecting speaker awareness into the model pretraining process, in order to increase the ability of the model to handle the bi-party turn-by-turn nature of dialogue modeling.

\chapter{Grounding dialogues on commonsense knowledge} 
\chaptermark{Grounding dialogues on commonsense knowledge}  
\label{ch:commonsense_chapter} 

Commonsense knowledge is considered a key part of human intelligence \cite{cambria2022senticnet}. Since dialogue systems are expected to respond to human utterances in an interesting and engaging way, commonsense knowledge has to be integrated into the model effectively.

In this chapter, we explain our effort on investigating the impact of providing commonsense knowledge about the concepts covered in the dialogue. In the retrieval-based scenario, we propose the Tri-LSTM model to jointly take into account message and commonsense for selecting an appropriate response. Our experiments suggest that the knowledge-augmented models are superior to their knowledge-free counterparts in automatic evaluation.







\section{Introduction}\label{sec:introduction}
By training on a large number of message-response pairs, most dialogue systems attempt to produce an appropriate response based solely on the message itself, without any memory module. In natural human conversation, however, people respond to each other's utterances in a meaningful way not only by paying attention to the latest utterance of the conversational partner itself, but also by recalling relevant information about the concepts covered in the utterance and integrating it into their responses. Such information may contain personal experience, recent events, commonsense knowledge and more (Figure \ref{fig:memory_module}). As a result, it's speculated that a conversational model with a ``memory look-up'' module can mimic human conversations more closely \cite{DBLP:journals/corr/GhazvininejadBC17,bordes2016learning}.
\begin{figure*}[h]
  \includegraphics[width=12cm, height=2cm]{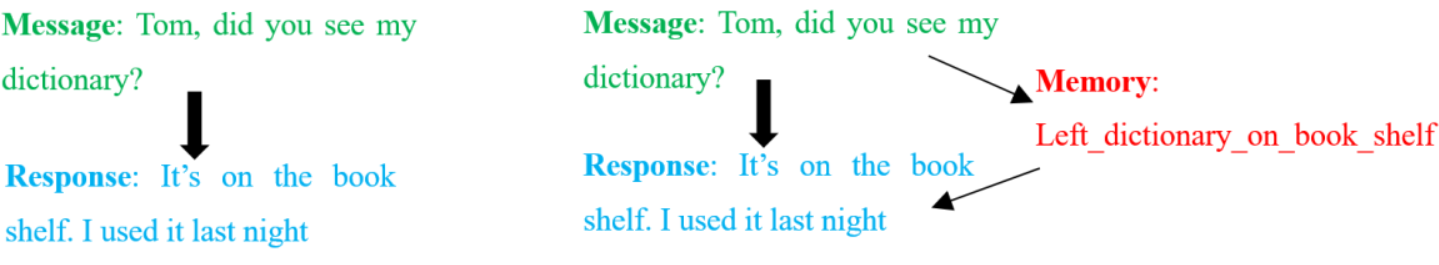}
  \centering
  \caption{Left: In traditional dialogue modeling, the response is determined solely by the message. (Arrows denote dependencies) Right: The responder recalls relevant information from memory about the message; memory and message jointly determine the response. As in the illustrated example, the responder model retrieves the event ``Left\_dictionary\_on\_book\_shelf'' from memory, which, along with the message, triggers a meaningful response.}
  \label{fig:memory_module}
\end{figure*}

In open-domain human-computer conversation, where the model is expected to respond to human utterances in an interesting and engaging way, commonsense knowledge has to be integrated into the model effectively. In artificial intelligence, commonsense knowledge is the set of background information that an individual is intended to know or assume and the ability to use it when appropriate \cite{minsoc,camcom,tratow}. Due to the vastness of such knowledge, we speculate that this goal might be better suited by employing an external memory module containing such knowledge than forcing the model to encode it in model parameters. Hence we investigate augmenting end-to-end dialogue systems with commonsense knowledge as external memory. 

Several commonsense knowledge bases have been constructed during the past decade, such as ConceptNet \cite{speer2012representing} and SenticNet \cite{cambria2014senticnet}. The aim is to give a foundation of real-world knowledge to a variety of AI applications. Typically a commonsense knowledge base can be seen as a \emph{semantic network} where \emph{concepts} are nodes in the graph and \emph{relations} are edges. Each $\textless concept1, relation, concept2 \textgreater$ triple is termed an \emph{assertion}. Based on the Open Mind Common Sense project \cite{singh2002open}, ConceptNet not only contains objective facts such as ``Paris is the capital of France'' that are constantly true, but also captures informal relations between common concepts that are part of everyday knowledge such as ``A dog is a pet''. This feature of ConceptNet is desirable for our purpose, because the ability to recognize the informal relations between common concepts is necessary in the open-domain conversation setting in this chapter.



\section{Proposed Approaches}\label{sec:model_description}

\label{headings}

\subsection{Task Definition}\label{sec:task_definition}

Our effort concentrated on integrating commonsense knowledge into retrieval-based conversational models as the first step, as they were easier to evaluate \cite{liu2016not,lowe2016evaluation} and generally took a lot less data to train before the dawn of large-scale pretrained models. We left the generation-based scenario to future work. 

\emph{Message} (\emph{context}) $x$ and \emph{response} $y$ are a sequence of tokens from vocabulary $V$. Given $x$ and a set of response candidates $[y_1,y_2,y_3,...,y_K]\in Y$, the model chooses the most appropriate response $\hat{y}$ according to:
\begin{eqnarray}
\hat{y}=\mathop{\arg\max}_{y\in{Y}}f(x,y),
\label{eq1}
\end{eqnarray}

%

where $f(x,y)$ is a scoring function measuring the ``compatibility'' of $x$ and $y$. 
The model is trained on $\textless message, response, label \textgreater$ triples with cross entropy loss, where $label$ is binary indicating whether the $\textless message, response \textgreater$ pair comes from real data or is randomly combined as a negative example. 

\subsection{Dual-LSTM Encoder}\label{duel_LSTM_encoder}

The Dual-LSTM encoder~\cite{lowe2015ubuntu} represents the message $x$ and response $y$ as fixed-size embeddings $\vec{x}$ and $\vec{y}$ with the last hidden states of the same LSTM. The compatibility function of the two is thus defined by:
\begin{eqnarray}
f(x,y) = \sigma(\vec{x}^{T}W\vec{y}),
\end{eqnarray}

where matrix $W \in \mathcal{R}^{D\times D}$ is learned during training. Dual-LSTM Encoder is used as both the baseline and the backbone model for our proposed model Tri-LSTM (Chapter \ref{sec:tri_lstm_encoder}).

\begin{figure*}[ht]
 \includegraphics[height=10.5cm]{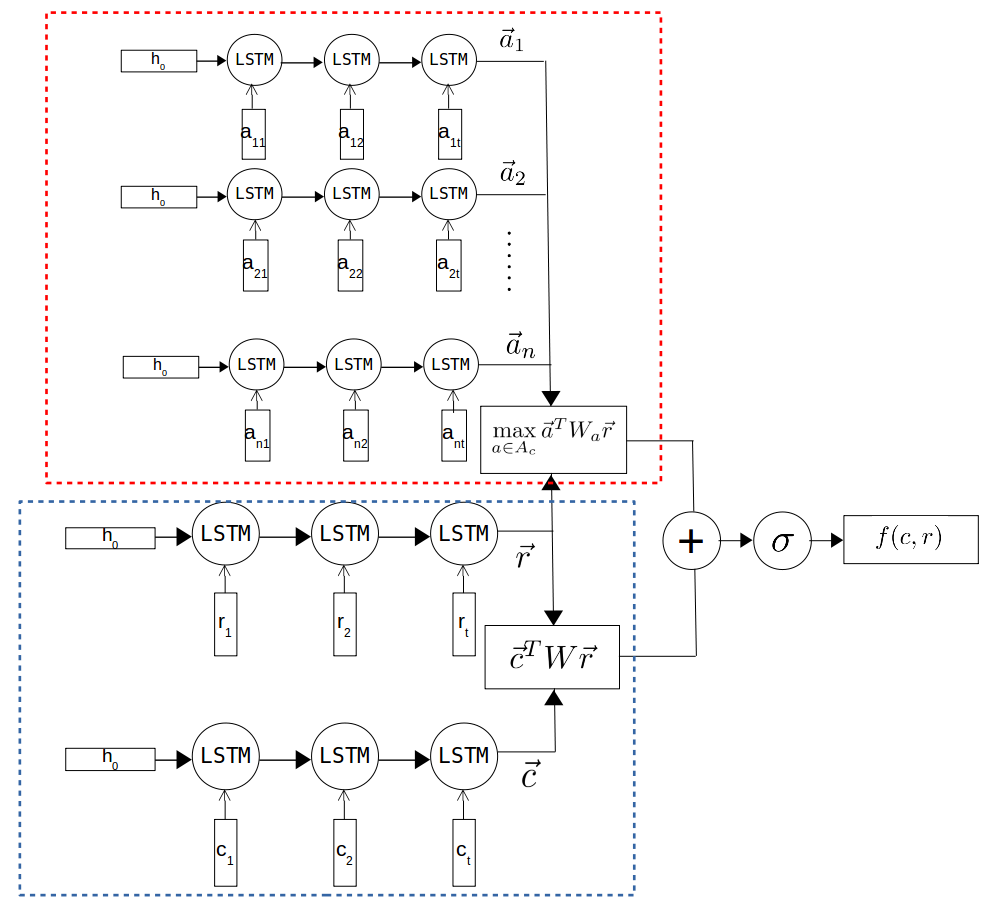}
 \centering
 \caption{Tri-LSTM encoder. We use LSTM to encode message, response and commonsense assertions. LSTM weights for message and response are tied. The lower box is equal to a Dual-LSTM encoder. The upper box is the memory module encoding all commonsense assertions.}
 \label{fig:triencoder}
\end{figure*}

\subsection{Commonsense Knowledge Retrieval}\label{sec:commonsense_knowledge_retrieval}

We assume that a commonsense knowledge base is composed of assertions $A$ about concepts $C$. Each assertion $a \in A$ takes the form of a triple $\textless c_1,r,c_2 \textgreater$, where $r \in R$ is a \emph{relation} between $c_1$ and $c_2$, such as \emph{IsA}, \emph{CapableOf}, etc. $c_1,c_2$ are concepts in $C$. The relation set $R$ is typically much smaller than $C$. $c$ can either be a single word (e.g., ``dog'' and ``book'') or a multi-word expression (e.g., ``take\_a\_stand'' and ``go\_shopping'').
We build a dictionary $H$ out of $A$ where every concept $c$ is a key and a list of all assertions in $A$ concerning $c$, i.e., $c=c_1$ or $c=c_2$, is the value.
Our goal is to retrieve commonsense knowledge about every concept covered in the message. 

\begin{figure*}[ht]
\includegraphics[height=8cm]{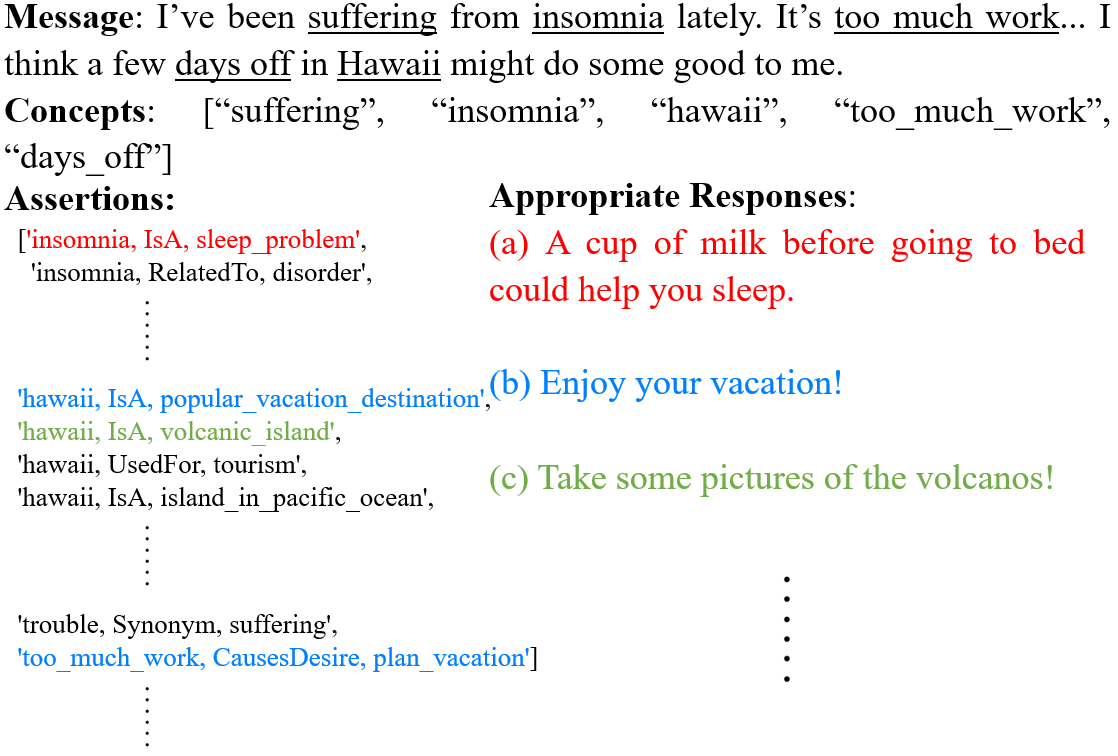}
 \centering
 \caption{In the illustrated case, five concepts are identified in the message. All assertions associated with the five concepts constitute $A_x$. We show three appropriate responses for this single message. Each of them is associated with (same color) only one or two commonsense assertions, which is a paradigm in open-domain conversation and provides ground for our max-pooling strategy. It is also possible that an appropriate response is not relevant to any of the common assertions in $A_x$ at all, in which case our method falls back to Dual-LSTM.}
 \label{fig:instance_csk}
\end{figure*}

We define $A_x$ as the set of commonsense assertions concerned with message $x$. To recover concepts in message $x$, we use simple $n$-gram matching ($n\leq N$)\footnote{More sophisticated methods such as $concept\ parser$~\cite{rajagopal2013graph} are also possible tools. Here, we chose n-gram for better speed and recall. $N$ is set to 5.}. Every $n$-gram in $c$ is considered a potential concept\footnote{For unigrams, we exclude a set of stopwords. Both the original version and stemmed version of every word are considered.}. If the $n$-gram is a key in $H$, the corresponding value, i.e., all assertions in $A$ concerning the concept, are added to $A_x$ (Figure~\ref{fig:instance_csk}).

\subsection{Tri-LSTM Encoder}\label{sec:tri_lstm_encoder}

Our main approach to integrating commonsense knowledge into the conversational model involves using another LSTM for encoding all assertions $a$ in $A_x$, as illustrated in Figure~\ref{fig:triencoder}. Each $a$, originally in the form of $\textless c_1,r,c_2 \textgreater$, is transformed into a sequence of tokens by chunking $c_1$, $c_2$, concepts which are potentially multi-word phrases, into $[c_{11},c_{12},c_{13} ...]$ and $[c_{21},c_{22},c_{23} ...]$. Thus, $a=[c_{11},c_{12},c_{13}, ...,r,c_{21},c_{22},c_{23}...]$. 

We add $R$ to vocabulary $V$, that is, each $r$ in $R$ will be treated like any regular word in $V$ during encoding. We decide not to use each concept $c$ as a unit for encoding $a$ because $C$ is typically too large ($>$1M).
$a$ is encoded as embedding representation $\vec{a}$ using another LSTM. Note that this encoding scheme is suitable for any natural utterances containing commonsense knowledge in addition to well-structured assertions. We define the \emph{match score} (compatibility score) of assertion $a$ and response $y$ as:
\begin{eqnarray}
m(a,y) = \vec{a}^{T}W_a\vec{y},
\end{eqnarray}


where $W_a \in \mathcal{R}^{D\times D}$ is learned during training. 
The number of commonsense assertions $A_x$ associated with a message is usually large ($>$100 in our experiment). 
We observe that in a lot of cases of open-domain conversation, response $y$ can be seen as triggered by certain perception of message $x$ defined by one or more assertions in $A_x$, as illustrated in Figure~\ref{fig:instance_csk}. For example, the word `Insomnia' in the message is related to the commonsense assertion `Insomnia, IsA, sleep$\_$problem'. The appropriate response (`go to bed') is then matched to `sleep$\_$problem'. Similarly, the word `Hawaii' in the message is related to the commonsense assertion `Hawaii, UsedFor, tourism'. The appropriate response (`enjoy vacation') is then matched based on `tourism'. 

Our assumption is that $A_x$ is helpful in selecting an appropriate response $y$. However, usually very few assertions in $A_x$ are related to a particular response $y$ in the open-domain setting. As a result, we define the \emph{match score} of $A_x$ and $y$ as\footnote{Here the symbol $m$ is overloaded with 2 meanings.}
\begin{eqnarray}
m(A_x,y)=\mathop{\max}_{a\in{A_x}} m(a,y),
\end{eqnarray}

%

that is, we only consider the commonsense assertion $a$ with the highest match score with $y$, as most of $A_x$ are not relevant to $y$. Incorporating $m(A_x,y)$ into the Dual-LSTM encoder, our Tri-LSTM encoder model is thus defined as:
\begin{eqnarray}
f(x,y) = \sigma(\vec{x}^{T}W\vec{y} + m(A_x,y)),
\end{eqnarray}
i.e., we use simple addition to supplement $x$ with $A_x$, without introducing a mechanism for any further interaction between $x$ and $A_x$. This simple approach is suitable for response selection and proves effective in practice. 

The intuition we are trying to capture here is that an appropriate response $y$ should not only be compatible with $x$, but also related to certain memory recall triggered by $x$ as captured by $m(A_x,y)$. In our case, the memory is commonsense knowledge about the world. In cases where $A_x = \emptyset$, i.e., no commonsense knowledge is recalled, $m(A_x,y)=0$ and the model degenerates to Dual-LSTM encoder.

\subsection{Comparison Approaches}\label{comparison_approaches}

%

\subsubsection{Supervised Word Embeddings}\label{supervised_word_embeddings}
We follow~\cite{bordes2016learning,dodge2015evaluating} and use supervised word embeddings as a baseline. Word embeddings are most well-known in the context of unsupervised training on raw text as in~\cite{mikolov2013efficient}, yet they can also be used to score message-response pairs. The embedding vectors are trained directly for this goal. In this setting, the ``compatibility'' function of $x$ and $y$ is defined as:
\begin{eqnarray}
f(x,y)=\vec{x}^T\vec{y}
\end{eqnarray}
In this setting, $\vec{x},\vec{y}$ are bag-of-words embeddings.
With retrieved commonsense assertions $A_x$, we embed each $a\in{A_x}$ to bag-of-words representation $\vec{a}$ and have:
\begin{eqnarray}
f(x,y)=\vec{x}^T\vec{y}+\mathop{\max}_{a\in{A_x}} \ \ \vec{a}^T\vec{y}.
\end{eqnarray}
This linear model differs from Tri-LSTM encoder in that it represents an utterance with its bag-of-words embedding instead of RNNs.

\subsubsection{Memory Networks}\label{memory_networks}
Memory networks~\cite{sukhbaatar2015end,weston2014memory} are a class of models that perform language understanding by incorporating a memory component. They perform attention over memory to retrieve all relevant information that may help with the task. In our dialogue modeling setting, we use $A_x$ as the memory component. Our implementation of memory networks, similar to~\cite{bordes2016learning,dodge2015evaluating}, differs from supervised word embeddings described above in only one aspect: how to treat multiple entries in memory.
In memory networks, output memory representation $\vec{o}=\sum_{i}p_i\vec{a}_i$, where $\vec{a}_i$ is the bag-of-words embedding of $a_i\in{A_x}$ and $p_i$ is the attention signal over memory $A_x$ calculated by $p_i=softmax(\vec{x}^T\vec{a_i})$. The ``compatibility'' function of $x$ and $y$ is defined as:
\begin{eqnarray}
f(x,y)=(\vec{x}+\vec{o})^T\vec{y}=\vec{x}^T\vec{y}+(\sum_{i}p_i\vec{a}_i)^T\vec{y}
\end{eqnarray}

In contrast to supervised word embeddings described above, attention over memory is determined by message $x$.

\section{Experiments and Results}\label{experiments}

\subsection{Twitter Dialogue Dataset}\label{twitter_dialogue_dataset}

1.4M Twitter \textless message, response$>$ pairs were used for our experiments. They were extracted over a 5-month period, from February through July in 2011. 1M Twitter \textless message, response$>$ pairs are used for training. With the original response as ground truth, we construct 1M \textless message, response, label=1$>$ triples as positive instances. Another 1M negative instances \textless message, response, label=0$>$ are constructed by replacing the ground truth response with a random response in the training set.

For tuning and evaluation, we use 20K \textless message, response$>$ pairs that constitute the validation set (10K) and test set (10K). They are selected by a criterion that encourages interestingness and relevance: both the message and response have to be at least 3 tokens long and contain at least one non-stopword. For every message, at least one concept has to be found in the commonsense knowledge base. For each instance, we collect another 9 random responses from elsewhere to constitute the response candidates. 

Preprocessing of the dataset includes normalizing hashtags, ``@User'', URLs, emoticons. Vocabulary $V$ is built out of the training set with 5 as minimum word frequency, containing 62535 words and an extra $\textless UNK \textgreater$ token representing all unknown words.

\subsection{ConceptNet}\label{conceptNet}
In our experiment, ConceptNet\footnote{https://conceptnet.io.} is used as the commonsense knowledge base. Preprocessing of this knowledge base involves removing assertions containing non-English characters or any word outside vocabulary $V$. 1.4M concepts remain. 0.8M concepts are unigrams, 0.43M are bi-grams and the other 0.17M are tri-grams or more. Each concept is associated with an average of 4.3 assertions. More than half of the concepts are associated with only one assertion.

An average of 2.8 concepts can be found in ConceptNet for each message in our Twitter Dialogue Dataset, yielding an average of 150 commonsense assertions (the size of $A_x$). Unsurprisingly, common concepts with more assertions associated are favored in actual human conversations.

It is worth noting that ConceptNet is noisy due to uncertainties in the constructing process, where 15.5\% of all assertions are considered ``false'' or ``vague'' by human evaluators~\cite{specon}. Our max-pooling strategy used in the Tri-LSTM encoder and supervised word embeddings is partly designed to alleviate this weakness. 

\subsection{Parameter Settings}\label{parameter_settings}

In all our models excluding term frequency--inverse document frequency (TF-IDF)~\cite{ramos2003using}, we initialize word embeddings with pretrained GloVe embedding vectors~\cite{pennington2014glove}. The size of hidden units in LSTM models is set to 256 and the word embedding dimension is 100. We use stochastic gradient descent (SGD) for optimizing with batch size of 64. We fixed training rate at 0.001.

\subsection{Results and Analysis}\label{results_and_analysis}
The main results for TF-IDF, word embeddings, memory networks and LSTM models are summarized in Table~\ref{tab:model_performance}. We observe that:

(1) LSTMs perform better at modeling dialogues than word embeddings on our dataset, as shown by the comparison between Tri-LSTM and word embeddings. 

(2) Integrating commonsense knowledge into conversational models boosts model performance, as Tri-LSTM outperforms Dual-LSTM by a certain margin. 

(3) Max-pooling over all commonsense assertions depending on response $y$ is a better method for utilizing commonsense knowledge than attention over memory in our setting, as demonstrated by the gain of performance of word embeddings over memory networks.

\begin{table*}[ht]
\centering
\footnotesize
\begin{tabular}{|c|c|c|c|c|c|c|}
\hline
Recall@$k$ & TF-IDF & \begin{tabular}[c]{@{}c@{}}Word \\ Embeddings$^*$\end{tabular} & \begin{tabular}[c]{@{}c@{}}Memory \\ Networks$^*$\end{tabular} & Dual-LSTM & Tri-LSTM$^*$   & Human \\ \hline
Recall@1 & 32.6\% & 73.5\% & 72.1\% & 73.6\%  & \textbf{77.5\%} & 87.0\% \\ \hline
Recall@2 & 47.3\% & 84.0\% & 83.6\% & 85.6\%  & \textbf{88.0\%} & -   \\ \hline
Recall@5 & 68.0\% & 95.5\% & 94.2\% & 95.9\%  & \textbf{96.6\%} & -   \\ \hline
\end{tabular}
\caption{Model evaluation. $^*$ indicates models with commonsense knowledge integrated. The TF-IDF model is trained following~\cite{lowe2015ubuntu}. The ``Recall@$k$'' method is used for evaluation~\cite{DBLP:journals/corr/LoweSNCP16}. The model is asked to rank a total of $N$ responses containing one positive response and $N-1$ negative responses ($N=10$ according to our test set). If the ranking of the positive response is not larger than $k$, Recall@$k$ is positive for that instance.}
\label{tab:model_performance}
\end{table*}

\begin{table*}[ht]
\centering
\begin{tabular}{|c|c|c|}
\hline
Instance & Message                                                                   & \begin{tabular}[c]{@{}c@{}}Response selected \\ by Dual-LSTM\end{tabular}                     \\ \hline
1    & \begin{tabular}[c]{@{}c@{}}i was helping my brother \\ with his chinese.\end{tabular}                            & did yoga help?                                                   \\ \hline
2    & \begin{tabular}[c]{@{}c@{}}bonjour madame, quoi de neuf.
\end{tabular}                                  & yeah me too!                                                   \\ \hline
3    & \begin{tabular}[c]{@{}c@{}}help what colour shoes can i \\ wear with my dress to the wedding?\end{tabular}                & very pale pink or black.$^\diamondsuit$                                              \\ \hline
4    & \begin{tabular}[c]{@{}c@{}}helping mum paint my bedroom.\end{tabular}                                  & 
\begin{tabular}[c]{@{}c@{}}what color are \\ you going for?$^\diamondsuit$\end{tabular} \\ \hline \hline
Instance & \begin{tabular}[c]{@{}c@{}}Response selected \\ by Tri-LSTM\end{tabular}                                  & \begin{tabular}[c]{@{}c@{}}Activated Assertion\\ (total size of $A_x$)\end{tabular}                  \\ \hline
1    & \begin{tabular}[c]{@{}c@{}}the language sounds interesting! \\ i really gotta learn it!$^\diamondsuit$\end{tabular}                  & \textit{\begin{tabular}[c]{@{}c@{}}chinese, IsA, \\ human\_language (755)\end{tabular}}               \\ \hline
2    & \begin{tabular}[c]{@{}c@{}}loool . you can stick with english ,\\ its all good unless you want\\ to improve your french.$^\diamondsuit$\\ \end{tabular} & \textit{\begin{tabular}[c]{@{}c@{}}bonjour, IsA, \\ hello\_in\_french (9)\end{tabular}}               \\ \hline
3    & very pale pink or black.$^\diamondsuit$                                                           & \textit{\begin{tabular}[c]{@{}c@{}}pink, RelatedTo, \\ colour (1570)\end{tabular}}                 \\ \hline
4    & \begin{tabular}[c]{@{}c@{}} what color are \\ you going for?$^\diamondsuit$\end{tabular}             & \textit{\begin{tabular}[c]{@{}c@{}}paint, RelatedTo, \\ household\_color (959)\end{tabular}}            \\ \hline
\end{tabular}
\caption{Case studies for the impact of commonsense assertions. ``Activated Assertion'' is the commonsense assertion entry in $A_x$ chosen by max-pooling. $\diamondsuit$ indicates correct selection. All 4 instances displayed are taken from the test set.}
\label{case_studies}
\end{table*}

We also analyze samples from the test set to gain an insight on how commonsense knowledge supplements the message itself in response selection by comparing the Tri-LSTM encoder and the Dual-LSTM encoder. 

As illustrated in Table~\ref{case_studies}, instances 1,2 represent cases where commonsense assertions as an external memory module provide certain clues that the model without one fails to capture. For example in instance 2, Tri-LSTM selects the response ``...improve your french'' to message ``bonjour madame'' based on a retrieved assertion ``$bonjour, IsA, hello\_in\_french$'', while Dual-LSTM selects an irrelevant response. 
Unsurprisingly, Dual-LSTM is also able to select the correct response in some cases where certain commonsense knowledge is necessary, as illustrated in instance 3. Both models select ``... pink or black'' in response to message ``...what color shoes...'', even though Dual-LSTM does not have access to a helpful assertion ``$pink, RelatedTo, color$''. 

Informally speaking, such cases suggest that to some extent, Dual-LSTM (a model with no external knowledge) is able to encode certain commonsense knowledge in model parameters (e.g., word embeddings) in an implicit way. 
In other cases, e.g., instance 4, the message itself is enough for the selection of the correct response, where both models do equally well.




\section{Related Work}\label{sec:related_work}

As an early research effort on developing dialogue systems that are conditioned on commonsense knowledge, our work was inspired by previous work along related directions. At the same time, we gladly notice following work that further explored the relationship between knowledge and dialogue systems. This section discusses both categories of related work.

{\large \textbf{Preceding work}}


We start with a discussion about the grounding works that our effort was based upon. The use of an external memory module in NLP tasks had received considerable attention, such as in question answering \cite{weston2015towards} and language modeling \cite{sukhbaatar2015end}. It had also been employed in dialogue modeling in several limited settings. With Memory Networks, \cite{dodge2015evaluating} used a set of fact triples about movies as long-term memory when modeling reddit dialogues, movie recommendation and factoid question answering. 
Similarly in a restaurant reservation setting, \cite{bordes2016learning} provided local restaurant information to the conversational model. Researchers had also proposed several methods to incorporate knowledge as external memory into the Seq2Seq framework. \cite{DBLP:journals/corr/XingWWLHZM16} incorporated the topic words of the message obtained from a pre-trained LDA model into the context vector through a joint attention mechanism. \cite{DBLP:journals/corr/GhazvininejadBC17} mined FoodSquare tips to be searched by an input message in the food domain and encoded such tips into the context vector through one-turn hop. The Tri-LSTM model we proposed in this chapter shares similarities with \cite{Lowe-unstructured-text-2015-nips-workshop}, which encoded unstructured textual knowledge with RNN. Our work distinguished itself from previous research in that we considered a large heterogeneous commonsense knowledge base in the open-domain retrieval-based dialog setting.

{\large \textbf{Following work}}

Next we discuss the research done by the community that followed our work. 

Recent effort on building dialogue systems usually drew inspiration from the success of large-scale pretraining and either fine-tuned their dialogue models using large-scale pretrained language models as backbones \cite{ham2020end} or used large-scale dialogue data for pretraining \cite{zhang2019dialogpt, adiwardana2020towards}. Such models are often hundreds or even thousands times as large as earlier models. Such drastic increase in model size begs the question: is it possible to implicitly learn to converse with commonsense knowledge by encoding it in the model parameters without explicitly referring to an external knowledge base? After all, large language models have been shown to be able to learn commonsense knowledge implicitly \cite{zhou2020evaluating}.

The answer seems to be ``no'' as of 2022, considering the available empirical evidence. Knowledge-augmented dialogue systems continue to out-perform their knowledge-free counterparts as discussed below.

Blenderbot2 \cite{komeili2021internet} improved upon its predecessor Blenderbot \cite{roller2020recipes} by incorporating a search engine. During conversation, the model can generate contextual internet search queries, read the results, and incorporate that information when responding to the user's messages. This important knowledge-grounding technique enables the model to stay up-to-date in an ever-changing world.

SeeKeR (Search engine→Knowledge→Response) \cite{shuster2022language} proposed to apply a single language model for the 3 modular tasks that ground dialogue responses on knowledge: search, generating knowledge, and generating a final response. 


LaMDA (Language Models for Dialog Applications) \cite{thoppilan2022lamda} took a step forward by coupling the dialogue system with a toolset containing an information retrieval system, a calculator, and a translator. The dialogue system is capable of consulting all 3 external information sources at will in real-time.

We note that search engines are very often much more up-to-date than dialogue data collected under restricted scenarios. Therefore, by introducing a real-time information retrieval engine to the dialogue system, the temporal generalization problem \cite{lazaridou2021pitfalls} is effectively alleviated for response generation. Essentially, instead of being constrained by the static training examples, the dialogue system can reflect the most recent version of human knowledge. 

\section{Chapter Summary and Future Prospects}\label{conclusions}

In this chapter, we emphasized the role of external knowledge in conversational models. In the open-domain dialogue setting, we illustrated our effort on experimenting with commonsense knowledge as external memory and proposed a method of using the LSTM to encode commonsense assertions to enhance response selection \footnote{The work in this chapter has been published in \cite{youaug}}. 

Although the gains presented by our new method were not spectacular according to recall@k, our view represented the first attempt at integrating a large commonsense knowledge base that potentially describes the external world into conversational models as a memory component. 

The massive research effort put forward by the community that followed our work helped make great advance towards the goal of building knowledge-grounded dialogue systems. However, the current state-of-the-art systems like LaMDA still falls short compared to humans in many ways. While external knowledge improves output groundedness, the model can still generate responses that do not \textit{accurately} reflect the contents of authoritative external sources.  While the system can generally respond to a simple question based on a single knowledge snippet, it often makes mistakes given a context that requires sophisticated reasoning \cite{yasunaga2021qa}. These issues call for further research in the future.





\chapter{Improving the computational efficiency of large-scale response retrieval} 
\chaptermark{Improving the computational efficiency of large-scale response retrieval}  
\label{ch:retrieval-chapter}

Large neural networks have been the state-of-the-art machine learning models in recent years. Yet sometimes they are very slow to make inference with. When they are used for dialogue systems, their slow inference speed can sometimes make real-time application impossible. This chapter focuses on our effort on addressing this issue in the context of dialogue systems.

Specifically, we note that strong retrieval-based dialogue systems that are based on large pre-constructed natural response candidate sets can produce diverse and controllable responses. However, a large candidate set could be computationally costly, as every response candidate needs to be paired with the input context for scoring and ranking. In this chapter, we propose methods that support fast and accurate response retrieval systems that can operate on large-scale response candidate sets. 

We utilize a computationally efficient dual encoding scheme in which contexts and responses are encoded into a sentence embedding space individually, where inner product is used for scoring. To boost accuracy, we adopt a knowledge distillation approach where a very strong yet computationally expensive joint encoding model is used to facilitate training our encoders. We then significantly boost the retrieval speed by adopting a learning-based candidate screening method that predicts a subset from the whole response candidate set. We show in the experiments that our model performs strongly in terms of retrieval accuracy and human evaluation. At the same time our retrieval speed is improved by orders of magnitude.

\section{Introduction}

\begin{figure}[h]
\centering
\includegraphics[width=0.7\textwidth]{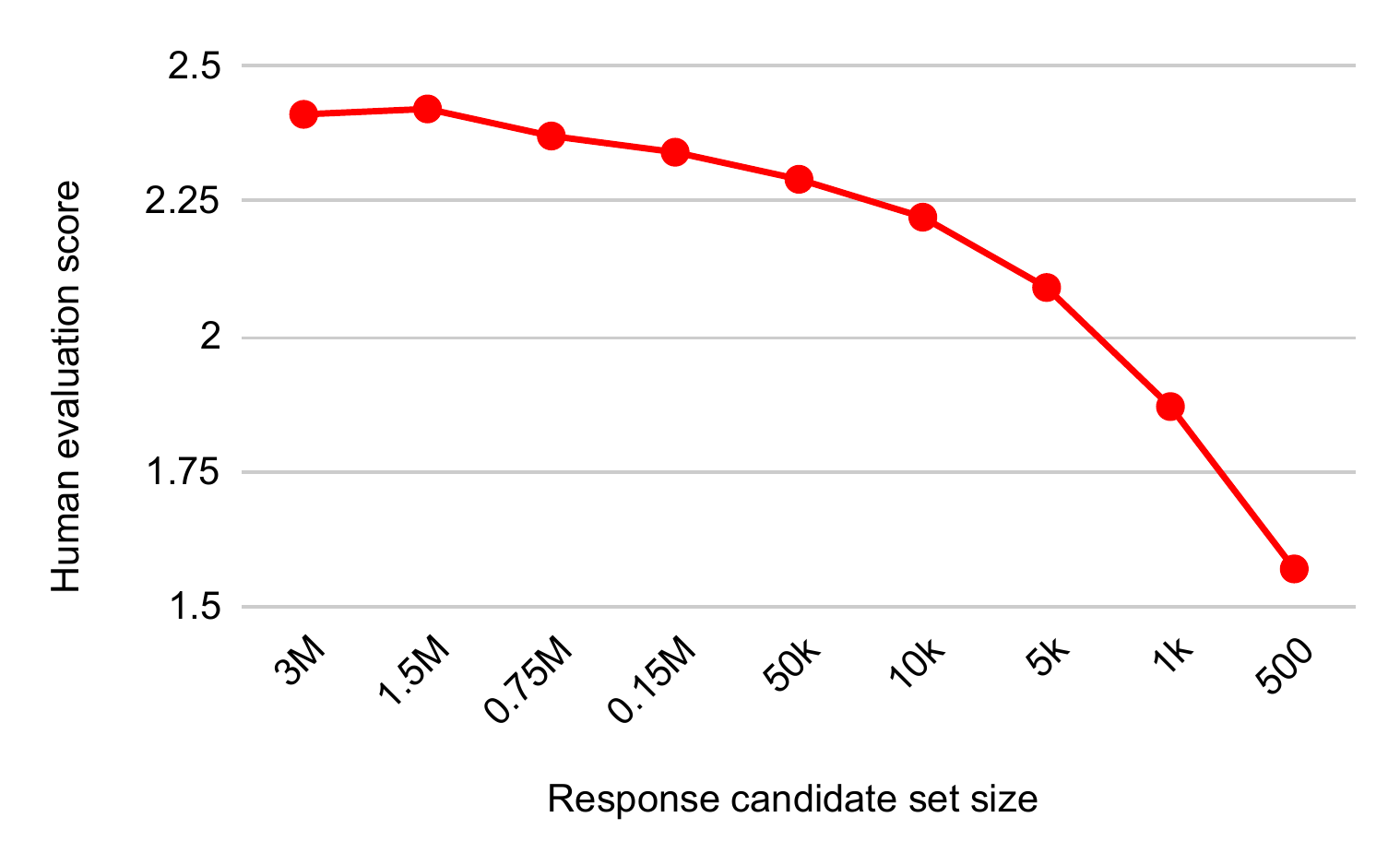}
\caption{Size of the response candidate set VS Response quality (human evaluation using the model of this paper's contribution as in Section \ref{sec:human_evaluation_on_response}).}\label{fig:set_size_vs_score_intro}
\end{figure}

Trained on a large set of natural context-response pairs, retrieval-based models attempt to select the most appropriate response from a response candidate set based on a scoring model that indicates the compatibility of a context and a response~\cite{lowe2015ubuntu,boussaha2019deep}. Compared to generation-based models, retrieval-based models have the advantage of being more controllable, since the responses come from a pre-constructed response set. Given a strong context-response pair scoring model, a retrieval-based system can be expected to produce long, interesting and diverse responses. The size and quality of the response candidate set matter a lot. Intuitively, a larger response set increases the possibility of finding a suitable response, especially for open-domain chitchat (Figure \ref{fig:set_size_vs_score_intro}). However, the size of the response set heavily affects retrieval speed, as the context needs to be paired with every response for scoring. Sometimes the retrieval time is linear with respect to the size of the candidate set, which makes some accurate but cumbersome retrieval networks practically infeasible~\cite{humeau2019polyencoders}. For example, the ``sentence pair classification'' setup in the BERT model \cite{devlin2018bert} can be directly applied to the retrieval-based dialogue scenario. The context and response are concatenated and fed into the transformer for scoring. However, pairing every response candidate with the context and forwarding them through the network during inference is computationally intractable for large response candidate sets. Another way is to encode the context
and response separately with 2 transformers of the same architecture and use their respective embeddings for scoring. Yet separate encoding loses accuracy.

Our work is driven by the idea of developing methods to achieve fast large-scale response retrieval while maintaining competitive accuracy. In our work, deep transformers are used to encode contexts and responses individually into a sentence embedding space, where inner product is used for scoring. This separate encoding scheme allows for much faster retrieval. To make up for the lost accuracy, we adopt a knowledge distillation approach where scores from the strong yet computationally expensive joint encoding model are used to facilitate the training process of our encoders. We further boost our retrieval speed by adopting a learning-based screening method that predicts a response candidate subset that the best response lies in based on the context. The subset is much smaller than the whole candidate set. Therefore, the retrieval time is drastically reduced.


The main contributions that we demonstrate in this chapter are:

(1)	We propose a fast large-scale response retrieval system based on knowledge distillation and deep transformer encoders. Knowledge distillation from a joint encoding model is performed for accuracy boosting.

(2)	A learning-based screening method is proposed for optimal response search based on maximum inner product in the embedding space, which enables very efficient response retrieval from large candidate sets.

(3)	Extensive experiments on both single-turn and multi-turn conversation settings show that our model performs favorably compared with strong retrieval-based baselines in terms of accuracy and speed trade-off. 

\section{Proposed Approaches}

\subsection{Overview}
\begin{figure}
\centering
\includegraphics[]{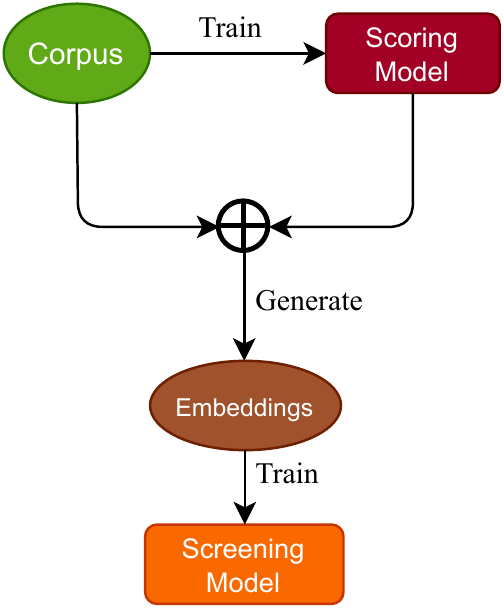}
\caption{Pipeline overview}\label{fig:model}
\end{figure}

A pipeline of the whole framework is shown in Figure~\ref{fig:model}. We have the scoring model (which contains the encoding model) that encodes the context and response into embeddings and assigns a matching score. It is trained on the corpus of context-response pairs. The screening model tries to predict a response subset given the context, it does so based on their embeddings. Thus the screening model is trained on embeddings produced by a fully trained scoring model run on the corpus. 




\subsection{Knowledge Distillation}

\begin{figure}[!tbp]
\centering
  \begin{subfigure}[b]{0.3\textwidth}
    \includegraphics[width=\textwidth]{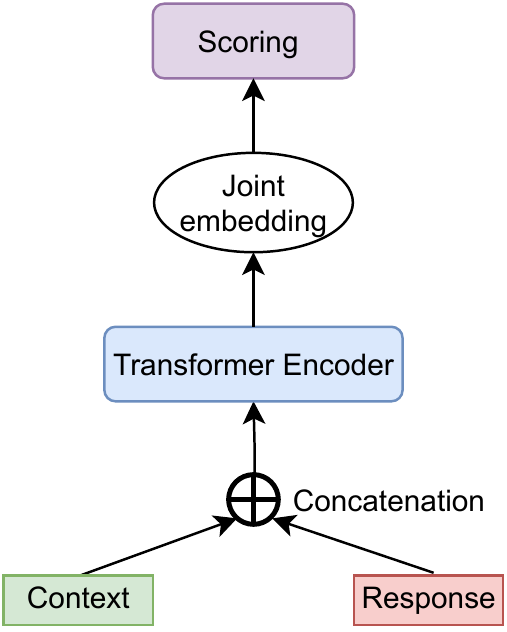}
    \caption{Cross-Encoder}
    \label{fig:cross_encoder}
  \end{subfigure}
  \medskip

  \begin{subfigure}[b]{0.4\textwidth}
    \includegraphics[width=\textwidth]{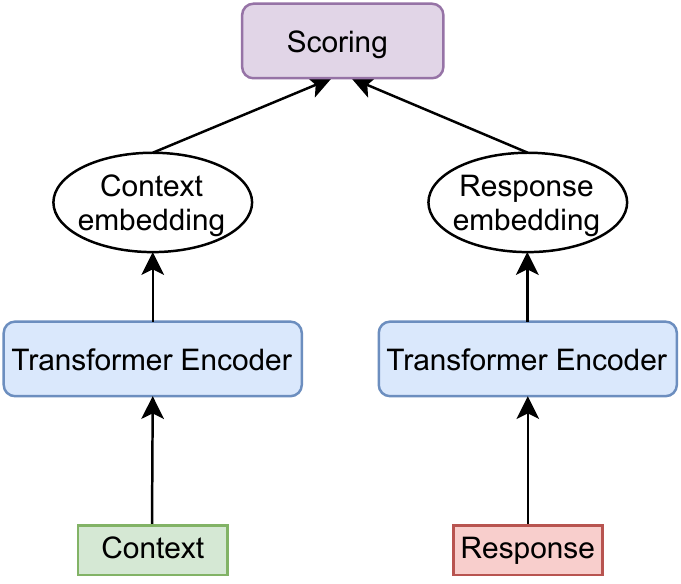}
    \caption{Dual-Encoder}
    \label{fig:dual_encoder}
  \end{subfigure}
  \caption{Two baseline encoding schemes that differ greatly in accuracy and speed.}
  \label{fig:two_baseline_models}
\end{figure}

One state-of-the-art method for response retrieval is to feed the concatenation of the context $c$ and the response $r$ through a deep transformer for joint encoding, termed the \textit{Cross-Encoder} (Figure~\ref{fig:cross_encoder}). The scoring function is defined as below.

\begin{eqnarray}
score_{\text{cross}} = \mathbf{FFN}(\mathbf{DT}([c; r])),
\label{eq_cross}
\end{eqnarray}

where $[ ; ]$ indicates concatenation and $\mathbf{FFN}$ is a feed forward network with Sigmoid activation. $\mathbf{FFN}$ reduces an embedding to a scalar score. $\mathbf{DT}$ stands for the deep transformer encoder, which encodes a word array into an embedding.

This encoding scheme allows the response to interact with input context in the deep transformer, which leads to high accuracy. However, the fact that during inference, the context needs to be paired with every response candidate for encoding and scoring, makes the retrieval process far too slow for large candidate sets. 

\begin{figure}
\centering
\includegraphics[]{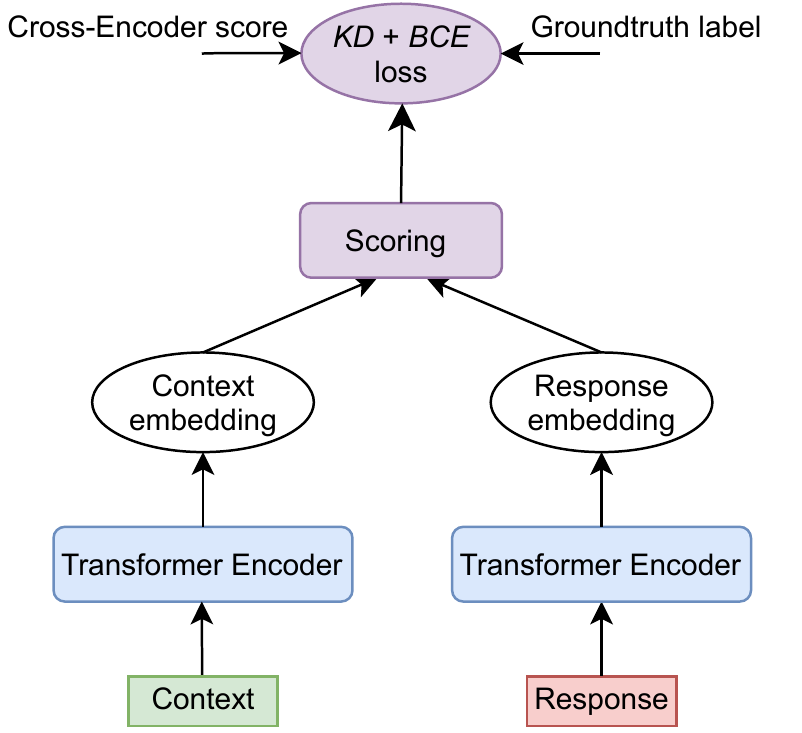}
\caption{Dual-Encoder with Knowledge Distillation}\label{fig:kd}
\end{figure}

In contrast, the \textit{Dual-Encoder} (Figure~\ref{fig:dual_encoder}) encodes the context and response separately, and uses the inner product between the two embeddings for scoring (Equations~\ref{c_dt}-\ref{eq_dual}). It loses accuracy as it does not allow direct interaction between the context and the response in the transformer.  However, with Dual-Encoders, we are able to cache the encoded candidate response embeddings, and reuse them for each new context. This results in significantly faster prediction.

\begin{eqnarray}
\mathbf{c} = \mathbf{DT}(c) \label{c_dt}\\
\mathbf{r} = \mathbf{DT}(r)
\label{r_dt}
\end{eqnarray}

\begin{eqnarray}
score_{\text{dual}} = sigmoid(\textbf{c}^T\textbf{r}),
\label{eq_dual}
\end{eqnarray}


where $\textbf{c}, \textbf{r} \in \mathbb{R}^D $ are the output embedding vectors of the deep transformer encoders. $D$ is the embedding dimension. To keep the Dual-Encoder's speed while boosting its accuracy, we adopt knowledge distillation. In our case, the scores given by the fully optimized Cross-Encoder are used to facilitate the training of the Dual-Encoder. Specifically, the loss used to train the new Dual-Encoder model (denoted as \textit{Dual-Encoder-KD}, Figure~\ref{fig:kd}) is the sum of the traditional cross entropy loss based on the ground-truth labels and the L2 loss between the Dual-Encoder-KD score and the Cross-Encoder score, as shown below.

\begin{align}
L_{\text{dual-kd}} = \beta(score_{\text{dual}} - score_{\text{cross}})^2 + \nonumber\\ BCE(score_{\text{dual}},label),
\label{eq_kd_l2}
\end{align}

where $\beta$ is a weighting coefficient, and $BCE$ is the traditional binary classification loss.

\subsection{Learning-based Candidate Screening}


Dual-Encoder allows encoding contexts and responses separately. Therefore, the retrieval time is drastically reduced by pre-computing and caching all the response candidate embeddings. According to Equation~\ref{eq_dual} and the monotonicity of the Sigmoid function, the best response in the response candidate set is found by

\begin{eqnarray}
\hat{r}=\mathop{\arg\max}_{r\in{R}}\textbf{c}^T\textbf{r},
\label{eq1}
\end{eqnarray}

where $R$ is the whole response candidate set. We further accelerate this maximum inner product search (MIPS) process. This algorithm uses a light-weight screening model to predict a much smaller set of candidate responses given the context, and then find the best response within that subset with vanilla calculation. 

This screening method has two sets of parameters - a set of $K$ context cluster centroids $\{\textbf{a}_1,..., \textbf{a}_K\} \in \mathbb{R}^D$, and the corresponding response subsets $\{\textbf{s}_1,..., \textbf{s}_K\} \in \{0, 1\}^N$, which are binary representations for which responses belong in the subset. $N$ is the total number of response candidates. A new context is assigned to a context cluster based on inner product with the centroids, and the corresponding response subset is the screening model's prediction. 

The probability of a context $\textbf{c}_i$ belonging to context cluster $k$ is modeled by
\begin{eqnarray}
\mu_{ik}=\dfrac{exp(\textbf{c}_i^T\textbf{a}_k)}{\sum_{l}exp(\textbf{c}_i^T\textbf{a}_l)}
\label{eq_belong}
\end{eqnarray}

The probability of retrieving $\textbf{r}_j$ for context $\textbf{c}_i$ is 

\begin{eqnarray}
p_{ij}=\sum_k\mu_{ik}\textbf{s}_k[j],
\label{eq_prob}
\end{eqnarray}

where $\textbf{s}_k[j] \in \{0, 1\}$ denotes the $j$th element of $\textbf{s}_k$. It indicates whether the $k$th cluster's subset includes response $\textbf{r}_j$. Essentially, equation \ref{eq_prob} describes $p_{ij}$ as the sum of the existence of $\textbf{r}_j$ across all subsets, weighted by cluster assignment.

The groundtruth best candidate label for $\textbf{c}_i$ and $\textbf{r}_j$ is $y_{ij} \in \{0, 1\}$, which indicates whether the best response for $\textbf{c}_i$ is $\textbf{r}_j$. 
Our loss is defined by

\begin{eqnarray}
l_{ij}=\begin{cases}
      \lambda p_{ij} & y_{ij}= 0 \\
      1-p_{ij} & y_{ij}= 1 
   \end{cases}
\label{eq_loss}
\end{eqnarray}

Another way to write it is:

\begin{eqnarray}
l_{ij}=\lambda p_{ij}(1-y_{ij}) + (1-p_{ij})y_{ij}
\label{eq_loss_2nd}
\end{eqnarray}

When $\textbf{r}_j$ is the best response for $\textbf{c}_i$, the model is punished by $1 - p_{ij}$. Otherwise it is punished by $p_{ij}$. $\lambda \in [0,1]$ is the balancing coefficient. It controls how much the screening model values speed vs. accuracy. Intuitively, the downside of including a single redundant candidate in the subset (e.g., $y_{ij} = 0$ and $p_{ij} = 1$) is much smaller than completely missing the ground-truth candidate (e.g., $y_{ij} = 1$ and $p_{ij} = 0$), thus $\lambda$ is set to be much smaller than 1. The overall optimization goal is thus the sum over all contexts and response candidates:

\begin{eqnarray}
\underset{\{{\textbf{a}_k}\}^K_{k=1}, \{{\textbf{s}_k}\}^K_{k=1}}{minimize}L = \sum_i\sum_{j}l_{ij}
\label{eq_objective}
\end{eqnarray}

To solve this optimization problem, we use alternating minimization. We optimize $\{{\textbf{a}_k}\}^K_{k=1}$ and $\{{\textbf{s}_k}\}^K_{k=1}$ alternatively while keeping the other one fixed. First, with $\{{\textbf{a}_k}\}^K_{k=1}$ fixed, $L$ can be expressed as
\begin{eqnarray}
L = \sum_k\sum_j\alpha_{kj}\textbf{s}_k[j] + \sum_i\sum_j y_{ij}
\label{eq_new_exp}
\end{eqnarray}
through Equation \ref{eq_prob} and \ref{eq_loss_2nd}, and $\alpha_{kj}$ is the coefficient for $\textbf{s}_k[j]$:

\begin{eqnarray}
\alpha_{kj} = \sum_i\mu_{ik}[\lambda - (\lambda+1)y_{ij}]
\label{eq_deduced_expression}
\end{eqnarray}

Remember $\textbf{s}_k[j] \in \{0, 1\}$. To minimize $L$, $\textbf{s}_k[j] $ is set to 0 if $\alpha_{kj} > 0$, otherwise it is set to 1.

Then we fix $\{{\textbf{s}_k}\}^K_{k=1}$ and update context cluster centroids $\{{\textbf{a}_k}\}^K_{k=1}$ (continuous, as opposed to the concrete $\{{\textbf{s}_k}\}^K_{k=1}$) using stochastic gradient descent (SGD) based on Equation~\ref{eq_objective}. We avoid the need to use the gumbel trick~\cite{jang2016categorical} as in~\cite{chen2018learning} by having defined the loss with probabilities. Following \cite{chen2018learning}, we initialize context cluster centroids $\{{\textbf{a}_k}\}^K_{k=1}$ with spherical k-means clustering. The overall learning process is given in Algorithm~\ref{algo}.






\begin{algorithm}
 \textbf{Input:} Context embeddings $\{{\textbf{c}_i}\}^M_{i=1}$, response candidate embeddings $\{{\textbf{r}_j}\}^N_{j=1}$, and labels $\textbf{Y} \in \{0, 1\}^{M \times N}$ indicating the best responses for each context given by exact inner product calculation according to Equation~\ref{eq1}.
 
 \textbf{Hyperparameter:}
 Number of clusters $K$, balancing coefficient $\lambda$, number of iterations $T$.
 
  \textbf{Output:} $K$ context cluster centroids $\{{\textbf{a}_k}\}^K_{k=1}$, $K$ response subsets $\{{\textbf{s}_k}\}^K_{k=1}$.
 
 \textbf{Training process:}
 Initialize $\{{\textbf{a}_k}\}^K_{k=1}$ by running spherical k-means clustering on $\{{\textbf{c}_i}\}^M_{i=1}$, Initialize  $\{{\textbf{s}_k}\}^K_{k=1}$ with 0's.
 
Execute the following procedures alternatively for $T$ iterations.

(1) Keep context cluster centroids  $\{{\textbf{a}_t}\}^K_{t=1}$ fixed and update response subsets  $\{{\textbf{s}_t}\}^K_{t=1}$  using Equation~\ref{eq_new_exp}.

(2) Keep response subsets  $\{{\textbf{s}_t}\}^K_{t=1}$  fixed and update context cluster centroids $\{{\textbf{a}_t}\}^K_{t=1}$ using SGD based on Equation~\ref{eq_objective}.
\medskip

\caption{Learning-based screening algorithm}
\label{algo}
\end{algorithm}

\section{Experiments and Results}

\subsection{Datasets}

Three datasets are used for our experiments. The first comes from Reddit online chat curated by~\cite{reddit}. It contains conversations from discussion threads on a variety of topics on the website Reddit. The second is the DailyDialog dataset~\cite{li2017dailydialog}, which contains human-written daily conversations that cover various topics. The third is the ConvAI2 dataset \cite{dinan2019second}, which is based on the PersonaChat dataset \cite{zhang2018personalizing}. The partitioning of the 3 datasets for the scoring models is shown in Table~\ref{tab:datastats}.

\begin{table*}[h!]
\centering
\begin{tabular}{|c|c|c|c|}
\hline
Dataset        & DailyDialog & ConvAI2    & Reddit  \\ \hline
No. Training   & 75000       & 131438     & 3300000 \\ \hline
No. Validation & 5000        & 7800       & 30000   \\ \hline
No. Testing    & 1000        & Unreleased & 10000   \\ \hline
\end{tabular}
\caption{The partitioning of the datasets for the \textit{scoring} models.}\label{tab:datastats}
\end{table*}

\begin{table*}[h!]
\footnotesize
\centering
\begin{tabular}{|c|c|c|c|c|}
\hline
\begin{tabular}[c]{@{}c@{}}Partitioning for \\ scoring models\end{tabular} & No. instances & \begin{tabular}[c]{@{}c@{}}Training contexts \\ for screening\end{tabular} & \begin{tabular}[c]{@{}c@{}}Testing contexts \\ for screening\end{tabular} & \begin{tabular}[c]{@{}c@{}}Screening candidate \\ responses\end{tabular} \\ \hline
Training                                                                   & 3300000       & yes                                                                        & no                                                                        & random 160k                                                              \\ \hline
Validation                                                                 & 30000         & yes                                                                        & no                                                                        & yes                                                                      \\ \hline
Testing                                                                    & 10000         & no                                                                         & yes                                                                       & yes                                                                      \\ \hline
\multicolumn{2}{|c|}{Total instances for screening}                                        & 3330000                                                                    & 10000                                                                     & 200000                                                                   \\ \hline
\end{tabular}
\caption{The origin of contexts and responses used for the \textit{screening} model. They are taken from the Reddit partitions for the \textit{scoring} models as in Table \ref{tab:datastats}. Yes/no indicates whether they come from the respective partitions.}\label{tab:screening_partitioning}
\end{table*}


Since our goal is to train a scoring model on whether or not a (context, response) pair is matched, we need negative pairs alongside the positive pairs that naturally exist in the dataset. For each positive (context, response) instance in the training and validation sets, we create a negative instance by replacing the ground-truth response with a random response in the training set.

For scoring-based retrieval models, a popular way to quantify their accuracy is to test how well they can identify the ground-truth response among distractors. Thus for Reddit and DailyDialog we build each instance in the test set by mixing the ground truth with 9 distractors. We then measure the frequency of the model scoring the ground-truth candidate higher than all other 9 distractor candidates, which is termed Recall@1/10. For ConvAI2, we report on the validation set  since the test set is not released. It is constructed for Recall@1/20 evaluation.

\subsection{Model Training Details}

We use the pre-trained BERT model based on~\cite{devlin2018bert} and~\cite{Wolf2019HuggingFacesTS} as our deep transformer encoder. Its output embedding length is 768. Our hyper-parameters are optimized with grid search. We train the Dual-LSTM (replacing the transformers in Dual-Encoder with LSTMs) with knowledge distillation from Cross-Encoder (\textit{Dual-LSTM-KD}) ~\cite{tang2019distilling} and re-implement the Poly-Encoder~\cite{humeau2019polyencoders} as additional baseline models as they were proposed as similar efforts to improve inference speed while maintaining good accuracy. The Poly-Encoder framework uses global-level self-attention on multiple context embeddings and the response embedding. We also run Poly-Encoder-KD as another reference, i.e., Poly-Encoder enhanced with knowledge distillation as in Dual-Encoder-KD.

For all the models on DailyDialog and Reddit, the best learning rate is searched for from options $\{5\mathrm{e}{-6}, 1\mathrm{e}{-5}, 5\mathrm{e}{-5}\}$. For the Poly-Encoder model, an extra hyperparameter “code length”~\cite{humeau2019polyencoders} is searched for from $\{16, 64, 128, 256\}$. For knowledge distillation, our weighting parameter $\beta$ comes from $\{0.2, 0.5, 1\}$. We use the Adam optimizer~\cite{kingma2014adam}. For ConvAI2, we report on the official validation set. Since this leaves us no practical validation set, the hyperparameters for ConvAI2 are simply the same as the best performing ones for Reddit.

We run our candidate screening model with different hyperparameters to find the best speed accuracy trade-off point. The number of clusters $K$ is from $\{10, 20, 50\}$ and the balancing coefficient $\lambda$ that balances accuracy and speed is from \\ $\{1\mathrm{e}{-5}, 5\mathrm{e}{-6}, 1\mathrm{e}{-6}, 5\mathrm{e}{-7}\}$. Since the purpose of the screening model is to improve speed for large-scale retrieval, we evaluate on the relatively large Reddit dataset. As our screening model requires supervised learning, it is most effective when the number of training contexts $M$ (conceptually similar to no. data points in multi-class classification) is significantly larger than the number of response candidates $N$ (similar to no. class categories). Therefore, we use 200k responses (including 160k from the scoring model training partition, and both validation and test partitions. See Table \ref{tab:screening_partitioning}) in the dataset as our response candidate set $\{{\textbf{r}_j}\}^N_{j=1}$. We leave 10k contexts (from the test partition) for testing and use all the rest in the dataset as the training contexts $\{{\textbf{c}_i}\}^M_{i=1}$ to train our screening model.



\subsection{Retrieval accuracy of the scoring models}\label{sec:recallk}

We conduct automatic evaluations on how well our scoring models can tell suitable responses apart from distractors. The retrieval accuracy results are shown in Table~\ref{tab:recallk}. The results suggest that on all three datasets, our Dual-Encoder-KD outperforms the baseline models with significant margins. Yet, it still falls short in comparison to its teacher model, that is Cross-Encoder. Poly-Encoder-KD and Dual-Encoder-KD perform very closely, in contrast to the stark difference between Poly-Encoder and Dual-Encoder. One possible explanation for this is that the supervision signal from Cross-Encoder is useful in a way that renders the more sophisticated architecture of Poly-Encoder less impactful.

Note that Recall@1/10 and Recall@1/20  can only test how reliable a scoring model scores context-response pairs, and cannot be used for evaluation, when a screening model is used. Because a screening model is only useful when a very large response candidate set is involved.



\begin{table*}[h!]
\centering
\begin{tabular}{|c|c|c|c|}
\hline
Dataset         & Reddit & DailyDialog & ConvAI2 \\ \hline
Dual-Encoder    & 0.814  & 0.710       & 0.802   \\ \hline
Cross-Encoder   & \textbf{0.882}  & \textbf{0.767}       & \textbf{0.831}   \\ \hline
Poly-Encoder    & 0.832  & 0.726       & 0.815   \\ \hline
Poly-Encoder-KD & 0.859  & 0.741       & 0.824   \\ \hline
Dual-LSTM-KD    & 0.810  & 0.703       & 0.801   \\ \hline
Dual-Encoder-KD & \textbf{0.857}  & \textbf{0.743}       & \textbf{0.822}   \\ \hline
\end{tabular}
\caption{Recall@1/10 results for Reddit and DailyDialog and Recall@1/20 results for ConvAI2.}\label{tab:recallk}
\end{table*}

\subsection{Accuracy and speed trade-off in candidate screening}\label{tradeoff}

For retrieving the best response from a large candidate set for a context, our candidate screening model drastically reduces the candidate set size at the cost of sometimes disregarding the best response. To quantitatively determine the performance of the model, we measure (1) \textit{speedup ratio}, defined as the expected ratio of the original candidate set size to the reduced candidate set size and (2) \textit{accuracy}, defined as the frequency of the reduced candidate set containing the best response in the original candidate set according to exact maximum inner product calculation.


We vary hyperparameters to achieve a good speedup ratio and accuracy trade-off. As shown in Table~\ref{tab:tradeoff}, in general, more speedup means less accuracy. As expected from Equation~\ref{eq_loss}, the higher $\lambda$ is, the more the model values speed. Our experiments suggest the more clusters there are, the smaller each subset is, and the faster and less accurate the model is. At $\lambda = 1\mathrm{e}{-6}$ and $K = 20$, we have a 5.14$\times$ speedup with a 1.3\% accuracy loss. This is the configuration we use in later experiments.

\begin{table*}[!htb]
    \begin{minipage}{0.4\linewidth}
      \centering
        \begin{tabular}{|c|c|c|c|c|}
        \hline
        \diagbox{$K$}{$\lambda$} & 1e-5   & 5e-6   & 1e-6   & 5e-7   \\ \hline
        10                     & 91.2\% & 95.2\% & 98.9\% & 99.1\% \\ \hline
        20                     & 90.9\% & 94.3\% & 98.7\% & 98.8\% \\ \hline
        50                     & 89.6\% & 92.7\% & 94.9\% & 95.3\% \\ \hline
        \end{tabular}
    \end{minipage}%
    \begin{minipage}{0.7\linewidth}
      \centering
        \begin{tabular}{|c|c|c|c|c|}
        \hline
        \diagbox{$K$}{$\lambda$}  & 1e-5  & 5e-6  & 1e-6  & 5e-7 \\ \hline
        10 & 12.79 & 8.79  & 4.36  & 3.31 \\ \hline
        20 & 14.70 & 10.33 & 5.14  & 4.92 \\ \hline
        50 & 29.18 & 21.07 & 10.31 & 9.27 \\ \hline
        \end{tabular}
    \end{minipage}
    \caption{Accuracy (left) and speedup ratio (right) given different hyperparameters. $K$ stands for the number of context clusters and $\lambda$ is the balancing coefficient.}\label{tab:tradeoff}
\end{table*}


\subsection{Retrieval speed in wall clock time}

We further measure the retrieval speed of the different models that have been discussed with wall clock time. Table~\ref{tab:retrieval_time} shows the average time it takes to retrieve a response from 200k candidates for a single context. Cross-Encoder is orders of magnitude slower than the rest. Dual-Encoder-KD outperforms Poly-Encoder. Our candidate screening model further increases the speed by about 5 times.

The retrieval process of Dual-Encoder-KD contains 2 steps: (a) Encode the context into an embedding using the encoder model and (b) Find the best response by running the context embedding against cached candidate response embeddings. In our experiment we find that step (a) takes less than 5\% as much time as step (b), due to the fact that the size of the candidate set is relatively large. With our screening model targeting time saving in step (b), the speedup achieved in wall-clock time by Dual-Encoder-KD-Screening compared to Dual-Encoder-KD is approximately the same as the ``speedup ratio'' in Section~\ref{tradeoff}.

\begin{table}[h!]
\centering
\begin{tabular}{|c|c|l|c|l|}
\hline
\multirow{2}{*}{Model}    & \multicolumn{4}{c|}{Retrieval time}                    \\ \cline{2-5} 
                          & \multicolumn{2}{c|}{CPU}   & \multicolumn{2}{c|}{GPU}  \\ \hline
Cross-Encoder             & \multicolumn{2}{c|}{90.5k} & \multicolumn{2}{c|}{8.3k} \\ \hline
Poly-Encoder              & \multicolumn{2}{c|}{90.2}  & \multicolumn{2}{c|}{12.3} \\ \hline
Dual-Encoder-KD           & \multicolumn{2}{c|}{22.5}  & \multicolumn{2}{c|}{6.9}  \\ \hline
Dual-Encoder-KD-Screening & \multicolumn{2}{c|}{4.1}   & \multicolumn{2}{c|}{1.7}  \\ \hline
\end{tabular}
\caption{Time in milliseconds to retrieve a response from 200k candidates. CPU computations were run on a 20 core Intel Xeon E5-2698v4. GPU computations were run on a single Tesla P100 with cuda 9.0. Dual (Poly)-Encoder and Dual (Poly)-Encoder-KD have the same speed as they share the same architecture.}
\label{tab:retrieval_time}
\end{table}











\subsection{Human evaluation on response quality}\label{sec:human_evaluation_on_response}


For subjective human evaluation, we resorted to Amazon Mechanical Turk (AMT) workers to directly score the quality of a single response produced by the system given the context. 200 random instances in the test set of Reddit were used. Each one was judged by 3 random judges that participated through AMT. Score levels and their meanings are [``3 - Very natural and appropriate'', ``2 - Somewhat relevant'', ``1 - Completely irrelevant'']. We treat the scores as numerical values and calculate their mean as a model's human evaluation response quality. Table~\ref{human_evaluation} shows the results. We see that human evaluation results for the retrieval-based models roughly align with Recall@1/10 results as in Section~\ref{sec:recallk}. Cross-Encoder performs the best by a small margin. It is followed by Dual-Encoder-KD, Poly-Encoder-KD and Dual-Encoder-KD-Screening, whose scores are very close to each other. Performing our screening method on Dual-Encoder-KD affects the score very marginally. Dual-Encoder-KD outperforms Dual-Encoder and Dual-LSTM-KD by a relatively large margin.


\begin{table}[h!]
\centering
\begin{tabular}{|c|c|}
\hline
Model               & Avg. score \\ \hline
Cross-Encoder       & \textbf{2.44}       \\ \hline
Poly-Encoder        & 2.35       \\ \hline
Dual-Encoder        & 2.29       \\ \hline
Dual-LSTM-KD     & 2.26       \\ \hline
Poly-Encoder-KD          & \textbf{2.43}       \\ \hline
Dual-Encoder-KD     & \textbf{2.39}       \\ \hline
Dual-Encoder-KD-Screening & \textbf{2.40}       \\ \hline
\end{tabular}
\caption{Human evaluation scores.}
\label{human_evaluation}
\end{table}

\section{Related Work} \label{sec:related_work}


The methods proposed in this chapter were inspired by previous works on knowledge distillation and maximum inner product search.

\subsection{Knowledge distillation in neural networks}
Distilling knowledge from a high-accuracy network into a low-accuracy network~\cite{hinton2015distilling} has proven to be an effective way to improve the accuracy of the latter. Traditionally the latter (student network) tries to mimic the former (teacher network) by minimizing a loss defined between the outputs of the two, in addition to the traditional loss based on groundtruth labels. As the multi-class output of the teacher network has higher entropy than the traditional one-hot labels, the student network has access to an information-rich similarity structure over data. The student network \textit{usually} has the advantage of being smaller, which makes training and inference faster. 

Research in various areas has shown the effectiveness of this approach. \cite{kim2016sequence} successfully boosted the inference speed of state-of-the-art machine translation networks by about 10 times with little loss in performance. It also found success in computer vision areas such as object detection~\cite{chen2017learning} and semantic segmentation~\cite{liu2019structured}.

Large-scale pretrained language models such as BERT are great teacher models. Recently there have been numerous efforts to distill knowledge from them and make new models that are smaller and faster. TinyBERT \cite{jiao2019tinybert} and BERT-PKD \cite{sun2019patient} distill knowledge from BERT through its embedding layers, hidden states and attention matrices into a smaller transformer-based model that is similar to BERT. \cite{tang2019distilling} showed that distilling knowledge from large-scale transformers into an LSTM makes the LSTM more competitive on sentence-level tasks. 


The works mentioned above view the student model as a neural network of smaller size. Our distillation approach, proposed for the specific scenario of large-scale response retrieval, is different. It is used on two models that contain the same original BERT architecture but with different representation formats of the input (context and response). For the teacher model, they are encoded together using one BERT, resulting in intractable inference and high accuracy. For the student model, two BERTs are used to encode context and response separately, resulting in faster inference but lower accuracy (Figure \ref{fig:two_baseline_models}). Experiments suggest that knowledge distillation also improves performance in our setting.

\subsection{Maximum inner product search}

Representing the context and response candidates in the embedding space separately is a time-efficient approach. With the matching criterion defined as inner product, the problem of finding the optimal response is reduced to a MIPS problem. Methods that accelerate MIPS have been studied in the setting of Neural Language Models. Neural Language Models consist of a softmax layer over a large vocabulary, which also poses as a MIPS problem during inference. Zhang et al.~\cite{zhang2018navigating} reduced MIPS to nearest neighbor search and showed a graph-based approach that performs well. The result of~\cite{zhang2018navigating} was later surpassed by~\cite{chen2018learning}. They proposed a learning based approach that assign vocabulary subsets to hidden state clusters, which inspired our work's screening model.

\section{Chapter Summary and Future Prospects}

In this chapter, we presented methods for a fast large-scale response retrieval model for human-computer interaction. Based on deep transformer encoders, we used knowledge distillation to leverage the learning power of a cumbersome joint encoding model to improve the performance of our fast individual encoders. Furthermore, to better handle large response candidate sets, we proposed a learning-based screening model that makes the retrieval process about 5 times faster with very little accuracy loss. Finally, we demonstrated a pipeline that performs strongly in terms of speed and quality trade-off compared with other retrieval-based models.

Apart from knowledge distillation, another promising direction for reducing the computational cost of neural networks is through sparse activation \cite{fedus2021switch}. Similar to the brain selectively activating different brain regions given different prompts, sparsely activating a large neural network would drastically reduce the computational cost.
\chapter{Summary} 
\chaptermark{Summary}  
\label{ch:Conclusion}


This thesis demonstrated our effort on innovating dialogue systems on multiple aspects, including (1) fusing different utilities, (2) fusing different modalities, (3) grounding dialogues on external knowledge and (4) improving the computational efficiency. Ultimately, an ideal dialogue system, equipped with useful knowledge bases, should be able to perform different functions seamlessly in a multi-modal process, and within a practical and affordable compute budget. These aspects for innovation for dialogue systems are also sought after for general AI paradigms.

Compared to other NLP tasks that focus on a specific aspect of language-related human intelligence, dialogue modeling can be seen as an all-in-one master task as it can naturally cover other tasks because of its very broad definition. Almost every NLP task can be converted to a dialogue modeling problem.

For example, question answering can be directly seen as a dialogue if one regards the question as the user message and the answer as the reply. Knowledge-grounded or multi-modal question answering can theoretically be covered by knowledge-grounded or multi-modal dialogue modeling. Even tasks that don't seem to be related to dialogues at all can be converted to dialogue modeling. For example, machine translation projects a sentence from its source language form to its target language form. By prepending ``please translate this sentence to English for me:'' to the source language sentence, the sentence pair becomes ``message + response'' like in dialogue systems. Dialogue modeling as a language generation task implicitly covers certain modular tasks. As an example, solving task-oriented dialogues implicitly requires solving named entity recognition and co-reference resolution.

Given such nature of dialogue systems - it may take on a very broad definition and cover almost all aspects of language-related human intelligence - it is no surprise that dialogue systems are a very challenging task. Yet every step further implies a considerable progress in AI. 


Looking into the future, we expect dialogue systems to continue to improve, very likely following general innovation in AI methodologies. For example, the following directions may be considered important for the long-term success of AI and thus also dialogue systems.

(1) Fast and easy generalization based on few-shot learning, zero-shot learning or meta learning. Compared to human intelligence, neural networks can be unbearably data-hungry. Thousands of carefully curated data points can be needed to learn to perform a new task. In contrast, humans require much less data (experience) because we can effectively utilize prior knowledge. Fortunately, recent research has began to shine light on this problem. For example, GPT3 \cite{brown2020language} demonstrated that a sufficiently large language model possesses remarkable few-shot or even zero-shot learning skills on a variety of tasks. \cite{reed2022generalist} took such realization and generalized it beyond text. Using a unified tokenization-based serialization scheme, their model ``Gato'' is pretrained on hundreds of tasks with different modalities and embodiments, such as image captioning, dialogues, and robotic control. Gato then displays few-shot learning capabilities in brand new tasks of similar categories.

(2) Several flaws of the state-of-the-art paradigms are not to be ignored regarding interpretablity and robustness. It has been noted that deep neural networks can be vulnerable to adversarial attacks. For example, minor phrase modification can easily deceive Google’s toxic comment detection systems \cite{hosseini2017deceiving}. In general, simple techniques such as word or pixel substitution that can't confuse humans can be used to fool neural networks \cite{dong2020towards}. Safety and controllability are also concerns especially for open-ended generation models. For example, Microsoft's dialogue system Tay learned to produce toxic responses in an online learning scenario \cite{wolf2017we}. In general, generation-based models may fail in unpredictable ways, thus restricting their usability to human-supervised scenarios \cite{pearce2021empirical}. Perhaps these problems are related to the black-box nature of neural networks, for which research on interpretability may be important.


\addtocontents{toc}{\vspace{0.8em}} 

\appendix 









\cleardoublepage    


\authorpublications{

\section*{Awards}

\begin{itemize}
    \item \href{https://cis.ieee.org/getting-involved/awards/2019-cis-award-recipients}{2021 IEEE CIM Outstanding Paper Award}
\end{itemize}


\section*{Journal Articles}
(Tom Young is the publication name for Yang Tianji.)
\begin{itemize}
    \item \textbf{Tom Young}, Vlad Pandelea, Soujanya Poria, and Erik Cambria. ``Dialogue systems with audio context.'' \emph{Neurocomputing} 388 (2020): 102-109.
    
    \item \textbf{Tom Young}, Devamanyu Hazarika, Soujanya Poria, and Erik Cambria. ``Recent trends in deep learning based natural language processing.'' \emph{IEEE Computational Intelligence Magazine} 13, no. 3 (2018): 55-75.
    
    \item Vlad Pandelea, Edoardo Ragusa, \textbf{Tom Young}, Paolo Gastaldo, and Erik Cambria. ``Toward hardware-aware deep-learning-based dialogue systems.'' \emph{Neural Computing and Applications} (2021): 1-12.
    
    \item Ni Jinjie, \textbf{Tom Young}, Vlad Pandelea, Fuzhao Xue, Vinay Adiga, and Erik Cambria. ``Recent advances in deep learning based dialogue systems: A systematic survey.'' under submission to Artificial Intelligence Review.
    
\end{itemize}

\section*{Conference Proceedings}

\begin{itemize}
  \item \textbf{Tom Young}, Erik Cambria, Iti Chaturvedi, Hao Zhou, Subham Biswas, and Minlie Huang. ``Augmenting end-to-end dialogue systems with commonsense knowledge.'' In \emph{Proceedings of the AAAI Conference on Artificial Intelligence}, vol. 32, no. 1. 2018.
  
  \item \textbf{Tom Young}, Frank Xing, Vlad Pandelea, Jinjie Ni, and Erik Cambria. ``Fusing task-oriented and open-domain dialogues in conversational agents.'' In \emph{Proceedings of the AAAI Conference on Artificial Intelligence}, vol. 36, no. 10, pp. 11622-11629. 2022.
  
  
  \item Ni Jinjie, Vlad Pandelea, \textbf{Tom Young}, Haicang Zhou, and Erik Cambria. ``HiTKG: Towards Goal-Oriented Conversations via Multi-Hierarchy Learning.'' In \emph{Proceedings of the AAAI Conference on Artificial Intelligence}, vol. 36, no. 10, pp. 11112-11120. 2022.
  
\end{itemize}

}

\backmatter


\label{Bibliography}
\setstretch{1}
\bibliographystyle{unsrtnat}
\bibliography{References/first_paper,References/second_paper,References/all_refs}

\end{document}